\pdfoutput=1

\documentclass[11pt]{article}

\usepackage[final]{acl}

\usepackage{times}
\usepackage{latexsym}

\usepackage[T1]{fontenc}

\usepackage[utf8]{inputenc} %
\usepackage{microtype} %
\usepackage{inconsolata} %
\usepackage{graphicx} %
\usepackage{xspace}
\usepackage{multirow}
\usepackage[normalem]{ulem}
\usepackage{enumitem}
\usepackage{booktabs}
\usepackage{amssymb}
\usepackage{colortbl} %
\usepackage{xcolor}   %
\usepackage{arydshln} %
\usepackage{titlesec} %

\titlespacing*{\section}{0pt}{*0.8}{*0.1}
\titlespacing*{\subsection}{0pt}{*0.8}{*0.1}
\titlespacing*{\subsubsection}{0pt}{*0.8}{*0.1}

\usepackage{amsfonts}       %
\usepackage{nicefrac}       %
\usepackage{amsmath}
\usepackage{amssymb}
\usepackage{bbm}
\usepackage{etoolbox}
\usepackage{csquotes}
\usepackage{makecell}
\usepackage{tabularray}
\usepackage{multirow}
\usepackage{pifont}%
\usepackage{ltablex}
\usepackage{wrapfig}
\usepackage{caption}
\usepackage{subcaption}
\usepackage{adjustbox}
\usepackage{longtable}
\usepackage{siunitx}

\usepackage{amsthm}

\DeclareMathOperator*{\argmax}{arg\,max}

\usepackage{tikz}
\usetikzlibrary{arrows}

\definecolor{mygreen1}{RGB}{22, 145, 36}
\definecolor{mypink1}{RGB}{255, 41, 134}

\newcommand{\deleted}[1]{{\color{mypink1}{\sout{#1}}}}
\newcommand{\deletedmay}[1]{{\color{mypink1}{\sout{#1}}}}
\newcommand{\specialdeleted}[1]{#1}

\renewcommand{\deleted}[1]{}

\renewcommand{\deletedmay}[1]{}
\renewcommand{\specialdeleted}[1]{}

\definecolor{myyellow1}{RGB}{240,165,5}
\definecolor{myred1}{RGB}{189, 44, 25}
\definecolor{myred2}{RGB}{240, 194, 194}
\definecolor{c_mistake}{RGB}{235, 174, 174}
\definecolor{c_fabrication}{RGB}{224, 144, 144}
\definecolor{mypurple1}{RGB}{206, 50, 217}
\definecolor{myorange1}{RGB}{255, 145, 0}

\newcommand{\addcite}[1]{{\textcolor{myred1}{\bf\sf[ADD CITE]}}\xspace}

\newcommand{\hide}[1]{}

\usepackage[normalem]{ulem}
\usepackage{enumitem}
\usepackage{xspace}

\newcommand{\mkt}{Model Knowledge Test\xspace}
\newcommand{\at}{Alignment Test\xspace}
\newcommand{\mks}{Model Knowledge Score\xspace}
\newcommand{\method}{{SHINE}\xspace}
\newcommand{\discovery}{{Entity Perturbation Impact}\xspace}

\makeatletter %

\renewcommand{\sectionautorefname}{\S\@gobble}
\renewcommand{\subsectionautorefname}{\S\@gobble}
\title{Probing LLM Hallucination from Within:\\
Perturbation-Driven Approach via Internal Knowledge}
\author{
    \textbf{Seongmin Lee\thanks{Work done during internship at JPMorganChase.}\textsuperscript{12}},
    \textbf{Hsiang Hsu\textsuperscript{2}},
    \textbf{Chun-Fu (Richard) Chen\textsuperscript{2}},
    \textbf{Duen Horng (Polo) Chau\textsuperscript{1}} \\
    \textsuperscript{1}Georgia Institute of Technology, \quad
    \textsuperscript{2}JPMorganChase Global Technology Applied Research
    \\
    \texttt{
    \href{seongmin@gatech.edu}{\color{black}seongmin@gatech.edu},
    \{\href{hsiang.hsu@jpmchase.com}{\color{black}hsiang.hsu},%
    \href{richard.cf.chen@jpmchase.com}{\color{black}richard.cf.chen\}@jpmchase.com},
    \href{polo@gatech.edu}{\color{black}polo@gatech.edu}
    }
}

\begin{document}
\maketitle

\begin{abstract}
LLM hallucination, where unfaithful  text is generated, presents a critical challenge for LLMs' practical applications. 
Current detection methods often resort to 
external knowledge, LLM fine-tuning, or supervised training with large hallucination-labeled datasets.
Moreover, these approaches do not distinguish between different types of hallucinations,
which is crucial for enhancing detection performance.
To address such limitations, we introduce \textbf{hallucination probing}, a new task that classifies LLM-generated text into three categories: \textit{aligned}, \textit{misaligned}, and \textit{fabricated}.
Driven by our novel discovery that perturbing key entities in prompts affects LLM's generation of these three types of text differently, we
propose \textbf{\method}, 
a novel %
hallucination probing method that does not require external knowledge, supervised training, or LLM fine-tuning. 
\method 
is effective in hallucination probing across three modern LLMs,
and achieves state-of-the-art performance in hallucination detection, outperforming seven competing methods across four datasets and four LLMs, underscoring the importance of probing for accurate detection.
 \end{abstract}

\begin{figure}[t]
    \centering
    \includegraphics[width=0.95\linewidth]{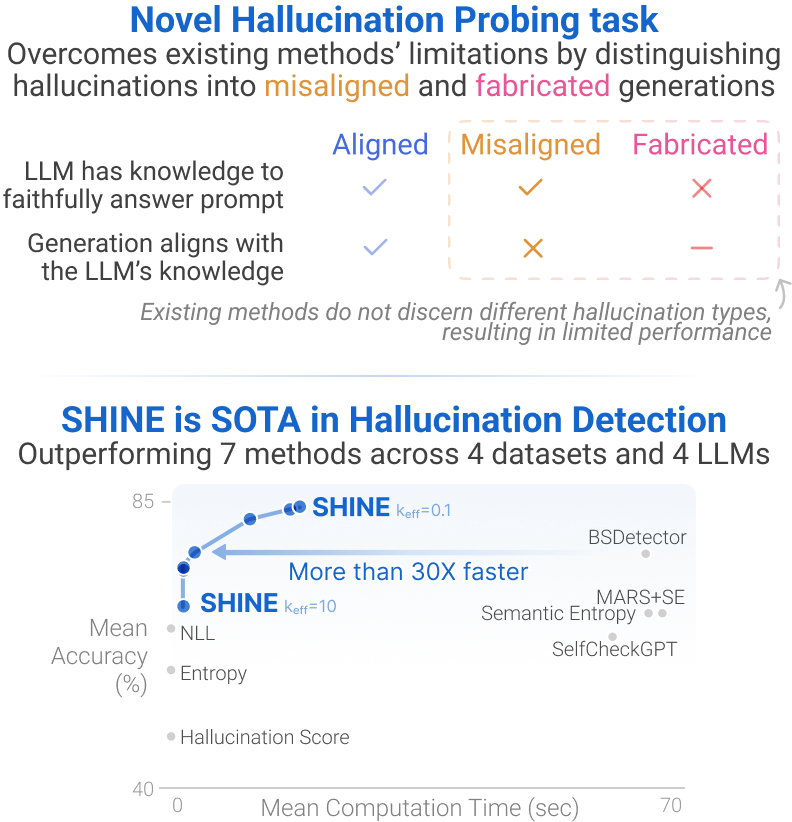}
    \vspace{-5pt}
    \caption{
    We introduce a novel \textbf{hallucination probing} task that distinguishes hallucination into \textit{misaligned} and \textit{fabricated} text. %
    \method, our novel hallucination probing method,
    achieves %
    \textbf{best} performance in hallucination detection, outperforming 7 methods across 4 datasets and 4 LLMs (\autoref{tab:hallucination_detection_performance_reasoning_dataset}, \ref{tab:hallucination_detection_performance_detection_dataset}).
    }
    \vspace{-10pt}
    \label{fig:crownjewel}
\end{figure}

\section{Introduction}
\label{sec:introduction}
Large language models (LLMs) excel at text 
generation~\cite{wu2023bloomberggpt,thirunavukarasu2023large}
but often produce hallucinations --- incorrect or unverifiable content --- posing significant risks for practical use~\cite{bohannon2023lawyer}.
Detecting such hallucinations is crucial yet challenging 
due to their plausible appearance~\cite{ji2023survey,xu2024hallucination} 
and diverse %
types~\cite{zhang2023siren,ji2023survey}.
One common detection approach involves
comparing outputs with external sources (e.g., Wikipedia)~\cite{lin2021truthfulqa,min2023factscore,tang2024minicheck}, but 
they fail when such 
sources are unavailable. %
Alternative approaches include fine-tuning LLMs to reject  questions likely to produce hallucinations~\cite{zhang2023r,xu2024rejection,li2024know} 
or training classifiers to detect them~\cite{azaria2023internal,chen2023hallucination,su2024unsupervised},
both requiring expensive supervised training on large %
labeled datasets.
To address these limitations, some recent efforts forgo  external knowledge or supervised training. These approaches include
checking consistency across multiple generations~\cite{manakul2023selfcheckgpt}, 
prompting LLMs to assess correctness~\cite{friel2023chainpoll,zhang2024self,chen2024quantifying},
or estimating LLM's uncertainty about its generation~\cite{zhang2023enhancing,farquhar2024detecting,kossen2024semantic,chen2024inside,yadkori2024believe}.

However, the methods above have a key limitation: they do not assess %
whether an LLM possesses sufficient knowledge to generate accurate responses ---
an essential factor in ensuring accurate detection.
Specifically,
consistency checks cannot identify hallucinations %
that the model repeatedly generates~\cite{slobodkin2023curious}, while 
prompting approaches may result in random guesses when the LLM lacks knowledge. 
Moreover, uncertainty-based methods assume all hallucinations stem from uncertainty, neglecting factors such as randomness in token sampling or beam search variability~\cite{stahlberg2019nmt,holtzman2019curious}. 
These limitations highlight the critical need %
to assess the LLM's knowledge and its alignment with generated content. We address this gap with the following key contributions:
\vspace{-5pt}
\begin{itemize}[topsep=0pt, itemsep=0mm, parsep=2pt, leftmargin=10pt]
    \item \textbf{A novel hallucination probing task} 
    that classifies LLM-generated text into 
    three categories ---
    \textit{aligned} (LLM has sufficient knowledge and responds faithfully), \textit{misaligned} (LLM has knowledge but 
    generates contradictions), 
    and \textit{fabricated} (LLM lacks relevant knowledge). %
    Our categorization refines hallucination types beyond existing work~\cite{huang2023survey,zhang2023siren} (\autoref{sec:analysis}), significantly boosting detection performance (\autoref{fig:crownjewel}).
    \item \textbf{\discovery{} discovery}, our key finding revealing that perturbing key entities in prompts affects LLM outputs differently depending on the type of hallucination present (\autoref{sec:analysis}).
    While prior studies have identified that hallucinations arise from errors while retrieving knowledge about the key entities~\cite{ferrando2024know},
    they did not leverage this for hallucination detection. 
    We systematically perturb key entities to assess their impact on generated tokens, 
    demonstrating how these perturbations can determine whether the generated tokens
    stem from the LLM's knowledge and align with it. This provides the foundation for distinguishing between aligned, misaligned, and fabricated responses for hallucination probing.  
    \item \textbf{\method, a novel 
    hallucination probing method} 
    leveraging our \textit{\discovery} discovery to predict whether an LLM possesses sufficient knowledge to generate a faithful response and whether the response aligns with the knowledge (\autoref{sec:method}, \autoref{fig:overview}).
    \method 
    operates without external knowledge, 
    supervised training, %
    or LLM fine-tuning. 
    \method stands for \textbf{S}ystematic \textbf{H}allucination \textbf{I}nspection with \textbf{N}oisy \textbf{E}ntity.
    \item \textbf{Extensive experiments demonstrating the superiority of our method} on four modern LLMs 
    (LLaMA2-13B-Chat, LLaMA3-8B-Instruct, Mistral-7B-Instruct, Qwen2.5-7B-Instruct) 
    across four datasets for diverse text generation tasks. %
    Our method effectively classifies aligned, misaligned, and fabricated text (\autoref{sec:hallucinaiton_reasoning}) and
    outperforms all existing algorithms on hallucination detection (\autoref{sec:hallucination_detection}), underscoring the importance of hallucination probing.
\end{itemize}
\section{Related Work}
\label{sec:related}
\textbf{Hallucination Probing.}
Previous studies have examined hallucination causes
by inspecting training data, algorithms, and inference process (Sec 3 of \cite{huang2023survey}). 
Key issues during inference include
contradictions with the LLM's knowledge --- arising from
its tendency to prioritize user preferences over accuracy, randomness in generation, and dependency on earlier tokens ---
and overconfident fabrication despite lacking knowledge
(Sec 4 of \cite{zhang2023siren}).
Based on this, we categorize hallucinations into two classes:
(1) misaligned, where the LLM has sufficient knowledge but produces contradictions, and
(2) fabricated, caused by the LLM's lack of knowledge.
Our work does not aim to attribute hallucination causes to model components~\cite{chen2025attributive}.

\smallskip
\noindent
\textbf{Hallucination Detection.}
Some approaches verify LLM-generated text using external sources or human evaluation, %
which can be unavailable or costly~\cite{lin2021truthfulqa,min2023factscore,tang2024minicheck}.
As an alternative,
researchers 
trained classifiers to detect %
hallucinations~\cite{chen2023hallucination,azaria2023internal,su2024unsupervised} or fine-tuned LLMs %
to avoid potentially hallucinated responses~\cite{zhang2023r,xu2024rejection,li2024know},
but these require training on large labeled datasets.
Prompting~\cite{friel2023chainpoll,li2023halueval,zhang2024self} or consistency checks~\cite{manakul2023selfcheckgpt,yadkori2024believe,yadkori2024mitigating,chen2024inside} 
may fail against confident fabrications~\cite{bengio2015scheduled,iqbal2022survey,zhang2023enhancing,kamath2024llm}. 
Uncertainty estimation methods~\cite{huang2023look,quevedo2024detecting,zhang2023enhancing,farquhar2024detecting,hou2024probabilistic,bakman2024mars,yaldiz2024not,liu2025enhancing}
often overlook hallucinations from randomness or token dependencies~\cite{stahlberg2019nmt,holtzman2019curious,zhang2024truthx}.
We propose a new direction to more accurately identify hallucinations  %
without any external knowledge, training, or impractical assumptions.

\section{Discovery: Entity Perturbation Impact for Hallucination Probing}
\label{sec:analysis}

Inspired by prior work that explores GPT-2's knowledge by perturbing entities in templated prompts~\cite{meng2022locating}, we ask: \textit{Is LLM-hallucinated text, ungrounded in model knowledge, as sensitive to perturbations as aligned text?} 
This drives our novel \discovery discovery, forming the basis of \method, our hallucination probing method (\autoref{sec:method}).
To explore this, we extend the analysis to 
modern LLMs, specifically, LLaMA2-13B-Chat-GPTQ~\cite{touvron2023llama} (henceforth, LLaMA2),
and more flexible prompts that ask factual questions about specific entities, which have been the primary focus of recent research~\cite{martino2023knowledge,sun2023head,ferrando2024know,huang2024can}.
We use the NEC benchmark~\cite{liu2024examining}, which includes prompts about real and fictional entities across six topics;
real-entity prompts yield aligned or misaligned responses, whereas fictional ones generate fabricated responses
(details in \autoref{sec:appendix:dataset:nec}).

Below, we outline an LLM's  inference process, then introduce our perturbation and evaluation method, which leads to our key discovery that perturbing key entities affects LLM's generation of aligned, misaligned, and fabricated responses differently, and ultimately gives rise to \method (\autoref{sec:method}).

\begin{figure}[t]
    \centering
    \includegraphics[width=0.95\linewidth]{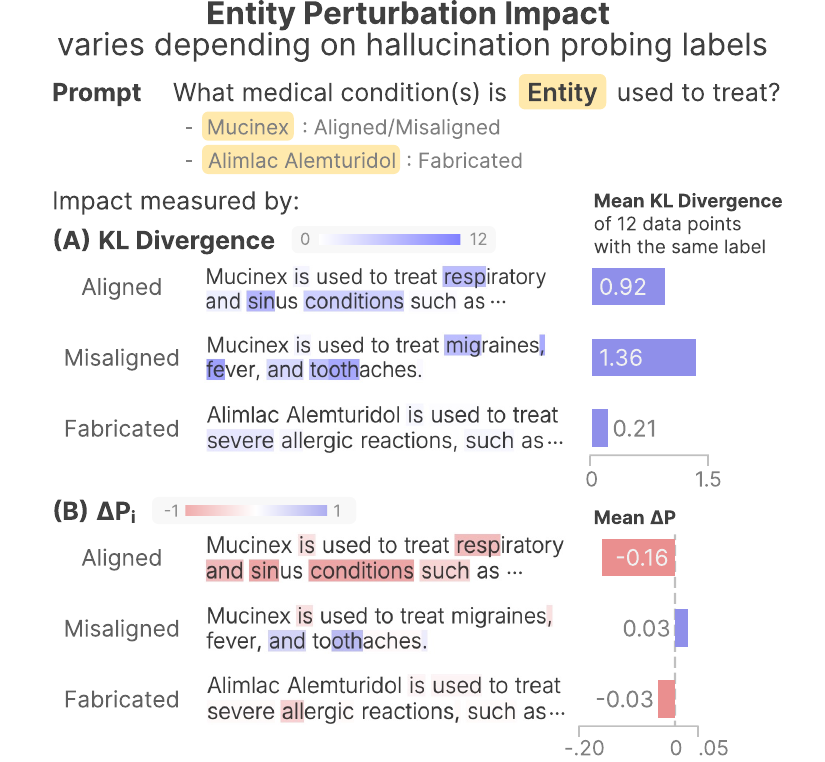}
    \vspace{-8pt}
    \caption{
    \discovery, measured by KL divergence and $\Delta P_i$, varies 
    across aligned, misaligned, and fabricated text.
    \textbf{(A)}
    Tokens in aligned and misaligned text, %
    strongly linked to the LLM's knowledge of the entity, 
    have high KL divergence, 
    while most tokens in fabricated text have near-zero values, indicating weak association with the LLM's knowledge. %
    The mean KL divergence of fabricated text is significantly lower than aligned and misaligned text.
    \textbf{(B)} While tokens in aligned and fabricated text generally show negative or near-zero $\Delta P_i$, 
    misaligned text has tokens with positive $\Delta P_i$, particularly those in the contradictory phrase,
    leading the mean $\Delta P$ of misaligned text to surpass that of aligned and fabricated text.
    }
    \vspace{-12pt}
    \label{fig:analysis_result}
\end{figure}

\noindent
\textbf{LLM inference without perturbation.}
For a prompt $P$ and text $G$, 
a tokenizer with a token set $\mathcal{T}$ splits $P$ into $M$ tokens $t_{1:M}=[t_1,\dots,t_M]$ and $G$ into $N$ tokens $t_{M+1:M+N}=[t_{M+1},\dots,t_{M+N}]$, 
where $t_i\in \mathcal{T}$.
Each token $t_i$ is mapped to a $d$-dimensional embedding vector $\mathbf{e}_i\in\mathbb{R}^d$ by a token embedding map.
An LLM $f$ takes the embedding vector sequence $\mathbf{e}_{1:M+N} = [\mathbf{e}_1,\dots,\mathbf{e}_{M+N}]$ as an input and computes the probability of each token %
appearing after each token position $i$; i.e., $f(\mathbf{e}_{1:M+N})=\mathbf{P}_{1:M+N}\in\mathbb{R}^{(M+N)\times\lvert\mathcal{T}\rvert},$ where $\mathbf{P}_i\in\mathbb{R}^{\lvert\mathcal{T}\rvert}$ and $\mathbf{P}_i(t)$ is the probability of the token $t$ to be generated at the position $i+1$.
We summarize these notations in \autoref{sec:appendix:notations}.

\noindent
\textbf{Entity Perturbation.}
For each prompt $P$, 
we manually identify the key entity $S$, tokenized as $t_{1:K}^{S}$.
We locate all token positions $I_S$ where $S$ appears in $P$ and the LLM's response $G$, %
i.e., $I_S=\{i\vert t_{i:i+K-1}=S\}$.
To perturb the embeddings of $S$, 
we add Gaussian noise $\boldsymbol\epsilon\sim\mathcal{N}(0,{(3\boldsymbol\sigma_0)}^2)\in\mathbb{R}^{K\times d}$ to all occurrences of $S$, where $\boldsymbol{\sigma}_0$ is the standard deviation of token embeddings in the NEC dataset~\cite{meng2022locating}; i.e., $\hat{\mathbf{e}}_{i:i+K-1}={\mathbf{e}}_{i:i+K-1}+\boldsymbol\epsilon$ for $i\in I_S$, while leaving other token embeddings unchanged.
Then, we input %
$\hat{\mathbf{e}}_{1:{M+N}}$ to the LLM $f$, obtaining a perturbed probability distribution $\hat{\mathbf{P}}_{1:M+N}=f(\hat{\mathbf{e}}_{1:{M+N}})\in\mathbb{R}^{(M+N)\times\lvert\mathcal{T}\rvert}$.
For each token position 
$i$ in $G$, 
we evaluate the \discovery using:
(1) Kullback-Leibler (KL) divergence between $\mathbf{P}_i$ and $\hat{\mathbf{P}}_i$ (i.e., $KL(\mathbf{P}_i\Vert\hat{\mathbf{P}}_i)$)
to capture global effects across the entire vocabulary
and
(2) change in the generation probability of token $t_i$ (i.e., $\Delta P_i=\hat{\mathbf{P}}_i(t_i)-{\mathbf{P}}_i(t_i)$), focusing on generated tokens.
To mitigate the impact of random Gaussian noise, we repeat this process 10 times with different random seeds. %

Our approach differs from methods that detect hallucinations via inconsistencies across generations.
While their inconsistencies mean that there is a token position $i$ with multiple inconsistent token $t$s with high generation probability $\mathbf{P}_i(t)$, 
we 
measure the distance between original and perturbed distributions ${\mathbf{P}}_i$ and $\hat{\mathbf{P}}_i$.
\noindent 
\textbf{Experiments.}
From the NEC dataset, we randomly sample two prompts for each topic for both existent and non-existent entities,
yielding 12 aligned, 12 misaligned, and 12 fabricated data points.
This analysis focuses on a small subset, %
as it requires manual identification of key entities;
results for full dataset
using our automated approach %
are in \autoref{sec:experiments}.
We also evaluate each text $G$'s \discovery 
by taking mean across all tokens in $G$
except those of the perturbed entity,
and refer to the results as
the KL divergence and $\Delta P$. %

\autoref{fig:analysis_result} demonstrates the
\discovery 
of aligned, misaligned, and fabricated text
(additional results in \autoref{sec:appendix:analysis}).
Fabricated text shows a much lower mean KL divergence (0.21) than aligned (0.92) and misaligned (1.36) text (\autoref{fig:analysis_result}A).
For example, %
in the aligned response to the prompt 
``\textit{What medical condition(s) is Mucinex used to treat?}'', %
tokens for \textit{respiratory} and \textit{sinus}, 
which are relevant to \textit{Mucinex},
have high KL divergence, %
whereas fabricated text's tokens exhibit near-zero KL divergence,
indicating their weak association with the LLM's knowledge.
Moreover, 
misaligned text contains tokens with positive $\Delta P_i$, with a mean $\Delta P$ of 0.03 --- higher than aligned (-0.16) and fabricated (-0.03) text, 
where most tokens have negative or near-zero $\Delta P_i$
(\autoref{fig:analysis_result}B). 
In the misaligned text about \textit{Mucinex},
tokens in the misaligned phrase \textit{fever, and toothaches} show positive $\Delta P_i$,
suggesting $\Delta P_i>0$ as a misalignment indicator.
\noindent
\textbf{Key Takeaways.}
Low \discovery measured by KL divergence 
strongly indicates fabricated text, 
as its tokens are weakly tied to  key entities.
In contrast, a rise in generation probability after perturbation ($\Delta P>0$) signals misalignment.
This is because aligned tokens are grounded in model knowledge of the key entities, 
so perturbation lowers their probability
($\Delta P<0$),
while tokens in fabricated text remains unaffected ($\Delta P\approx0)$.
In \autoref{sec:method}, we introduce \method, a hallucination probing method that leverages \discovery to distinguish between aligned, misaligned, and fabricated text. %

\vspace{-3pt}
\begin{figure*}[tb]
    \centering
    \includegraphics[width=0.95\linewidth]{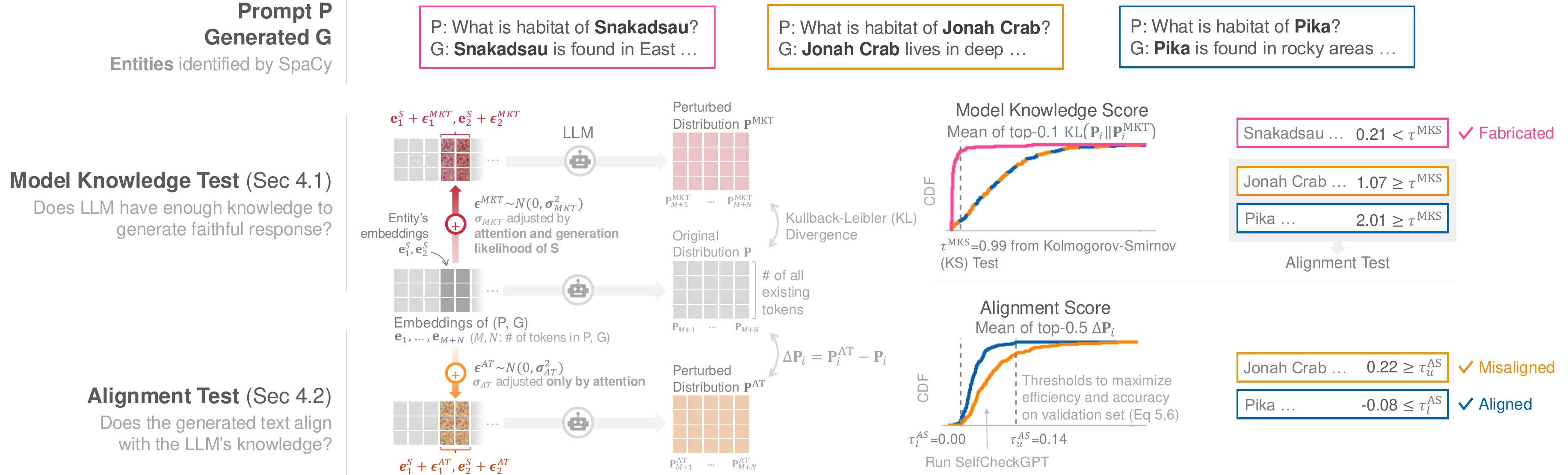}
    \vspace{-3pt}
    \caption{
    \method classifies LLM-generated text into aligned, misaligned, and fabricated 
    through two steps: \mkt (\autoref{sec:mkt}) and Alignment Test (\autoref{sec:at}). 
    The \mkt identifies whether the LLM has sufficient knowledge to generate a faithful response by perturbing entities in the prompt and %
    evaluating the impact; fabricated text is differentiated at this stage. For the data that the LLM has enough knowledge, we conduct the \at to examine whether the text aligns with the LLM's knowledge.
    }
    \vspace{-10pt}
    \label{fig:overview}
\end{figure*}

\section{\method: Hallucination Probing Method}
\label{sec:method}
Leveraging our \discovery discovery (\autoref{sec:analysis}),
we develop a two-stage workflow for hallucination probing, consisting of \textit{Model Knowledge Test} and \textit{Alignment Test} (\autoref{fig:overview}).
The \mkt first assesses whether the LLM has enough knowledge to generate faithful response for the prompt and distinguishes fabricated text from other two types (\autoref{sec:mkt}). 
Then, the Alignment Test examines whether the generated text aligns with the LLM's knowledge, classifying it as either aligned or misaligned (\autoref{sec:at}).

\subsection{Model Knowledge Test}
\label{sec:mkt}

\mkt builds on %
our discovery that fabricated text exhibits much lower \discovery (KL divergence) than aligned or misaligned text.
To systematically identify key entities, we detect all entities in a prompt and perturb them
with varying strength, determined by attention they receive.
The \mkt involves four steps: 
identifying entity, determining perturbation strength, perturbing entity embeddings, and measuring perturbation impact.
\smallskip
\noindent
\textbf{Step 1. Entity Identification.}
\label{sec:subject_identification}
Using the named entity and noun chunk extraction of SpaCy~\cite{spacy2},
\mkt extracts entities $S_1,\dots,S_L$ in $P$ (details in \autoref{sec:appendix:entity_identification}).
\noindent
\textbf{Step 2. Perturbation Strength Computation.}
To prioritize influential entities,
we apply stronger perturbations to entities with higher attention. 
To calculate the attention %
$att_l\in\mathbb{R}^+$ of each entity $S_l$~\cite{tu2021keywordmap},
we concatenate 
tokens in $P$ and $G$ into $t_{1:M+N}$, feed the token sequence into the LLM, and sum the attention values 
received by the tokens of $S_l$ while generating $t_{M+1:M+N}$.
We then scale $att_l$ into $a_l\in[0,1]$:
\begin{equation}
\vspace{-1pt}
\small
a_l={\exp{(k_{att}\cdot att_l)}\over{\exp{(k_{att}\cdot\max{(att_1,\dots,att_L)})}}},    
\vspace{-1pt}
\end{equation}
where $k_{att}\ge0$ %
controls the impact of attention values;
$k_{att}=0$ weighs all entities equally, 
while larger values emphasize highly attended entities.
Since LLMs tend to fabricate for 
entities with low frequency in the training data~\cite{mallen2023trust,kandpal2023large,min2023factscore,ji2023towards}, 
we further adjust perturbation strength 
using the generation likelihood of each entity $S_l$,
which serves as a good proxy for frequency~\cite{hartmann2023sok}.
Specifically, we use
the negative log-likelihood, scaled by token position to consider later tokens' importance. 
For an entity $S_l$, tokenized as $t_{1:K_l}^{S_l}$, 
the perturbation strength $w_l\in\mathbb{R}^+$ is calculated as
\begin{equation}
\vspace{-2pt}
\small 
    w_l=a_l\cdot\left(1-{1\over K_l}\sum_{k=1}^{K_l}{\sqrt{k-1}\log{Pr(t_k^{S_l}\vert t_{1:k-1}^{S_l})}}\right)^{-1}.
    \label{eq:fam_score}
\end{equation}
\noindent
\textbf{Step 3. Entity Perturbation.}
We perturb $L$ entities by adding noise to their token embeddings.
For entity $S_l=t_{1:K_l}^{S_l}$,
we identify all token positions $I_{S_l}$ where it appears in $(P,G)$:
$I_{S_l}=\{i\vert t_{i:i+K_l-1}=S_l\}$.
We add Gaussian noise 
$\boldsymbol{\epsilon}_l^{MKT} \sim \mathcal{N}(0, (\boldsymbol{\sigma}_l^{MKT})^2)\in\mathbb{R}^{K_l\times d}$
to all $i\in I_{S_l}$, 
i.e., ${\mathbf{e}}^{MKT}_{i:i+K_l-1}$ = $\mathbf{e}_{i:i+K_l-1} + \boldsymbol{\epsilon}_l^{MKT}$, 
while leaving other tokens unchanged;
the standard deviation 
$\boldsymbol{\sigma}_l^{MKT}=w_l\cdot\boldsymbol{\sigma}$, where $\boldsymbol{\sigma}$ is a hyperparameter (details in \autoref{sec:experiments}).
We input the perturbed embeddings ${\mathbf{e}}^{MKT}_{1:M+N}$ to the LLM 
to compute the perturbed probability distribution ${{\mathbf{P}}^{MKT}}=f({\mathbf{e}}_{1:M+N}^{MKT}) \in \mathbb{R}^{(M+N)\times \vert \mathcal{T} \vert}$. 
\noindent
\textbf{Step 4. \mks Evaluation.} 
We compute KL divergence between $\mathbf{P}$ and ${{\mathbf{P}}}^{MKT}$ at each token position $i$ in the generated text $G$ and 
refer to it as \mks at the position $i$, which is denoted as 
$MKS_i=KL(\mathbf{P}_i\Vert{\mathbf{P}}^{MKT}_i)$.
As the dependency on the previous tokens can reduce the KL divergence values of some tokens (details in \autoref{sec:appendix:analysis}),
we evaluate $MKS$, the \mks of $G$, by taking the mean of top-$p^{MKS}$ values of $MKS_i$ for $i=M+1,\dots,M+N$ ($0<p^{MKS}\le1$):
\vspace{-5pt}
\begin{equation}
\small
    MKS={ %
    {{1\over \lceil p^{MKS}\cdot N\rceil}
    {\sum_{i=1}^{\lceil p^{MKS}\cdot N\rceil}{MKS_i^{sort}}
    }}},
\vspace{-2pt}
\end{equation}
where $MKS_1^{sort}\ge\dots\ge MKS_N^{sort}$ is a sorted permutation of $MKS_{M+1},\dots,MKS_{M+N}$. %
We repeat this 10 times with different random seeds. 
If the \mks 
is lower than a threshold, we classify the text as fabricated; otherwise, we proceed to the Alignment Test (\autoref{sec:at}).
We determine the threshold $\tau^{MKS}$ using the Kolmogorov-Smirnov (KS) test~\cite{massey1951kolmogorov} on the validation set: $\tau^{MKS}=\argmax_{\tau\in[0,\infty)}{(\mathbf{F}_f^{MKS}(\tau)-\mathbf{F}_{a,m}^{MKS}(\tau))}$,
where $\mathbf{F}_f^{MKS}$ is the cumulative probability of fabricated data's \mks and $\mathbf{F}_{a,m}^{MKS}$ is that of aligned or misaligned data.
\subsection{Alignment Test}
\label{sec:at}

After ensuring the LLM has enough knowledge about $(P,G)$ via \mkt, 
we check if the text $G$ aligns with that knowledge.
Specifically,
we add noise 
$\boldsymbol{\epsilon}^{AT}_l\sim\mathcal{N}(0,(\boldsymbol{\sigma}_l^{AT})^{2})$
to the embeddings of each entity $S_l$ to compute the perturbed embeddings ${\mathbf{e}}_{1:M+N}^{AT}$;
since \mkt has already verified that the model is knowledgeable about $S_l$, 
we do not use the likelihood but adjust $\boldsymbol\sigma^{AT}_l$ only with $a_l$ as $\boldsymbol\sigma^{AT}_l=a_l\cdot\boldsymbol\sigma$, where $\boldsymbol\sigma$ is the same hyperparameter used in \mkt.
We then compute the perturbed probability distribution  ${\mathbf{P}}^{AT}=f({\mathbf{e}}_{1:M+N}^{AT})\in\mathbb{R}^{(M+N)\times\lvert\mathcal{T}\rvert}$ and
$\Delta P_i={\mathbf{P}}_i^{AT}(t_i)-\mathbf{P}_i(t_i)$ for $i=M+1,\dots,M+N$.
Since some tokens in the misaligned text can still be aligned with model knowledge, while all tokens in the aligned text should be aligned, we 
evaluate Alignment Score (AS) by taking the mean of top-$p^{AS}$ $\Delta P_i$ values ($0< p^{AS}\le 1$):
\begin{equation}
\small
    AS={
    {1\over {\lceil p^{AS}\cdot N\rceil}}
    {
    \sum_{i=1}^{\lceil p^{AS}\cdot N\rceil}{\Delta{P}_i^{sort}}
    }},
\end{equation}
where $\Delta P_1^{sort}\ge\dots\ge\Delta P_N^{sort}$ is a permutation of $\Delta{P}_{M+1},\dots,\Delta{P}_{M+N}$ sorted in descending order.

While large negative and large positive $AS$ values 
indicate %
alignment and misalignment, respectively, $AS\approx0$ can imply either. %
To handle near-zero $AS$,
we use
SelfCheckGPT~\cite{manakul2023selfcheckgpt}, which takes some computational cost 
but effectively differentiates text about which a model has enough knowledge.
We reduce the computational cost of SelfCheckGPT while achieving high accuracy
by running it only on data with near-zero $AS$.
Such data is identified by setting lower and upper thresholds, $\tau_{l}^{AS}$ and  $\tau_{u}^{AS}$, and selecting data within these thresholds; 
data points with AS lower than $\tau_l^{AS}$ and higher than $\tau_u^{AS}$ are directly classified as \textit{aligned} and  \textit{misaligned}, respectively.
The thresholds are determined to maximize correct predictions and minimize incorrect predictions on the validation set as 
\begin{equation}
\small
    \tau_l^{AS}=\arg\max_x{\left({
    {1+\mathbf{F}_{a}^{AS}(x)}
    \over
    {1+\mathbf{F}_{m}^{AS}(x)}/k_{eff}}
    \right)}
    \label{eq:as_lower_th}    
\end{equation}
\begin{equation}
\small
    \tau_u^{AS}=\arg\max_x{\left({
    {1+(1-\mathbf{F}_{m}^{AS}(x))}
    \over
    {1+(1-\mathbf{F}_{a}^{AS}(x))}/k_{eff}}
    \right)},
    \label{eq:as_upper_th}
\end{equation}
where $\mathbf{F}_{a}^{AS}$ and $\mathbf{F}_{m}^{AS}$ are the cumulative probabilities of the Alignment Score for aligned and misaligned data, %
respectively, and $k_{eff}$ is a hyperparameter that controls efficiency; larger $k_{eff}$ reduces penalization for misclassifications, allowing more data to be classified by $AS$; \autoref{sec:efficiency} further discusses how $k_{eff}$ balances the efficiency and performance.

\section{Experiments}
\label{sec:experiments}
This section demonstrates \method's effectiveness in hallucination probing (\autoref{sec:hallucinaiton_reasoning}) and detection (\autoref{sec:hallucination_detection}), highlighting its superiority over existing algorithms and the importance of probing. %
Then, we show how our method improves efficiency while maintaining high performance (\autoref{sec:efficiency}).

\subsection{Hallucination Probing}
\label{sec:hallucinaiton_reasoning}

We demonstrate \method's effectiveness on three  LLMs:
LLaMA2-13B-Chat-GPTQ~\cite{touvron2023llama},
LLaMA3-8B-Instruct~\cite{dubey2024llama},
and Mistral-7B-Instruct v0.3~\cite{jiang2023mistral},
henceforth referred to as LLaMA2, LLaMA3, and Mistral.
We use 
\texttt{en\_core\_web\_sm} model for SpaCy, %
LLM temperature of 1.0,
$\boldsymbol{\sigma}=10\boldsymbol{\sigma}_0$,  where $\boldsymbol{\sigma}_0$ is the standard deviation of token embeddings of each dataset,
$k_{att}$ of 0.1, and $k_{eff}$ of 0.1.
\subsubsection{Dataset}
\label{sec:hallucination_reasoning_dataset}
Our evaluation requires datasets of prompts, LLM-generated responses, and labels (aligned, misaligned or fabricated). 
Since existing datasets only provide binary labels (hallucinated or not), we create two new datasets---NEC and Biography---featuring trinary labels, using existing datasets~\cite{liu2024examining,min2023factscore}. 
The \textbf{NEC dataset} contains questions on existent and non-existent concepts and LLM-generated responses, with %
359 responses in each category (aligned, misaligned, fabricated) for LLaMA2, 359 for LLaMA3, and 476 for Mistral.
The \textbf{Biography dataset} contains real and fake biographies, with %
109 aligned, 109 misaligned, and 129 fabricated  for LLaMA2; 
174 aligned, 174 misaligned, and 174 fabricated for LLaMA3;
and 128 aligned, 128 misaligned, and 128 fabricated for Mistral.
Each dataset is split evenly into validation and test sets within each label.
Dataset construction details can be found in \autoref{sec:appendix:dataset}.

\subsubsection{Results}

We evaluate the \mks and Alignment Score by visualizing their cumulative distribution functions (CDF) on the validation set of the NEC dataset for the LLaMA2 model (\autoref{fig:overview}). %
The distributions of the \mks  for fabricated and non-fabricated text (i.e., aligned and misaligned) are substantially distinct, demonstrating the score's effectiveness in detecting fabrication. 
Specifically, the KS statistic
($\max_{\tau\in[0,\infty)}{\mathbf{F}_f^{MKS}(\tau)-\mathbf{F}^{MKS}_{a,m}(\tau)}$) is 0.82 at $\tau^{MKS}$ of 0.99 (p-value 1.53e-80).\footnote{$\tau^{MKS}$ is highly consistent aacross datasets, with values of 0.99 on NEC and 1.01 on the biography for LLaMA2.} %
Alignment Score distributions
for aligned and misaligned data also show a clear gap, enabling us to 
classify data with Alignment Score below $\tau_l^{AS}=0.00$ as aligned and above $\tau_u^{AS}=0.14$ as misaligned, 
and use SelfCheckGPT for the remaining data.
\begin{table}
\centering
\caption{Confusion matrix showing \method{}'s hallucination probing performance (\%) 
on LLaMA2, LLaMA3, and Mistral
across two datasets: NEC / biography.}
\label{tab:hallucination_reasoning_performance}
\begin{adjustbox}{max width=\linewidth}
\begin{tabular}{@{}cllccc@{}}
\toprule
& & & \multicolumn{3}{c}{Actual} \\ 
\cmidrule{4-6}
& & & Aligned  & Misaligned & Fabricated \\
\midrule
\multirow{3}{*}{LLaMA2} & 
\multirow{3}{*}{\rotatebox[origin=c]{90}{\small Predicted}} & Aligned & 
\textbf{75.53} / \textbf{81.82} & 13.53 / 5.45 & 9.89 / 4.62 \\
& & Misaligned &
11.17 / 9.09 & \textbf{73.53} / \textbf{87.27} & 1.65 / 9.23 \\
& & Fabricated  & 
13.30 / 9.09 & 12.94 / 7.27 & \textbf{88.46} / \textbf{86.15} \\
\midrule
\multirow{3}{*}{LLaMA3} & 
\multirow{3}{*}{\rotatebox[origin=c]{90}{\small Predicted}} & Aligned & 
\textbf{72.93} / \textbf{83.91} & 20.77 / 6.90 & 6.04 / 1.15 \\
& & Misaligned &
14.92 / 8.05 & \textbf{68.31} / \textbf{75.86} & 8.79 / 1.15 \\
& & Fabricated  & 
12.15 / 8.05 & 10.93 / 17.24 & \textbf{85.16} / \textbf{97.70} \\
\midrule
\multirow{3}{*}{Mistral} & 
\multirow{3}{*}{\rotatebox[origin=c]{90}{\small Predicted}} & Aligned & 
\textbf{73.44} / \textbf{89.06} & 11.62 / 3.12 & 12.45 / 1.56 \\
& & Misaligned &
5.81 / 4.69 & \textbf{65.15} / \textbf{87.50} & 4.15 / 3.12 \\
& & Fabricated  & 
20.75 / 6.25 & 23.24 / 9.38 & \textbf{83.40} / \textbf{95.31} \\
\bottomrule
\end{tabular}
\end{adjustbox}
\end{table}
\autoref{tab:hallucination_reasoning_performance} shows the confusion matrices of \method on the NEC and biography datasets across the LLaMA2, LLaMA3, and Mistral models.
Since no existing methods differentiate text into these three types, we cannot directly compare our approach with others. 
Overall, \method effectively differentiates the three types across both datasets and all three LLMs.
For LLaMA2, the \mkt correctly detects 88.46\% of fabricated data points in the NEC dataset and 86.15\% in the biography dataset.
The \at further differentiates aligned and misaligned text,
correctly classifying 75.53\% of aligned and 73.53\% of misaligned data points in the NEC dataset and 81.82\% of aligned and 87.27\% of misaligned data points in the biography dataset.
Similarly, for LLaMA3 and Mistral, \method consistently classifies most data points correctly, demonstrating its robust effectiveness in hallucination probing.
\subsection{Hallucination Detection}
\label{sec:hallucination_detection}

\begin{table*}[th]
    \centering
    \caption{\small
    \method{} (in \textbf{bold}) outperforms all existing approaches in hallucination detection (\%), especially for fabricated text, across the NEC / Biography datasets.
    For a fair comparison, we adapt \method's predictions to binary by combining its \underline{misaligned} and \underline{fabricated} classes into the conventional \textit{hallucinated} class, and mapping its \underline{aligned} class to the \textit{faithful} class.
    }
    \label{tab:hallucination_detection_performance_reasoning_dataset}
    \vspace{-4pt}
    \begin{adjustbox}{max width=\linewidth}
    \footnotesize
    \renewcommand{\arraystretch}{0.8}
    \begin{tabular}{ll*{4}{c}}
        \toprule 
        & & & Aligned  & Misaligned & Fabricated \\ 
        \cmidrule(l{15pt}r{15pt}){4-4}\cmidrule(l{10pt}r{10pt}){5-6}
        & & Overall & \textit{Faithful} & \multicolumn{2}{c}{\textit{Hallucinated}} \\ 
        \midrule
        \multirow{8}{*}{LLaMA2} & \textbf{\method (Ours)} & \textbf{83.69} / \textbf{90.58} & 73.40 / 81.82 & 86.47 / 94.55 & 91.21 / 95.38 \\
        & SelfCheckGPT & 63.58 / 76.92 & 86.70 / 81.82 & 86.47 / 98.18 & 17.58 / 50.77 \\
        & BSDetector & 76.69 / 73.80 & 85.64 / 70.91 & 82.35 / 98.18 & 62.09 / 52.31 \\
        & NLL & 65.00 / 57.58 & 82.98 / 89.09 & 90.59 / 83.64 & 21.43 / 0.00 \\
        & Entropy & 58.37 / 55.06 & 67.55 / 98.18 & 72.94 / 65.45 & 34.62 / 1.54 \\
        & Hallucination Score & 48.15 / 68.02 & 93.62 / 89.09 & 37.65 / 87.27 & 13.19 / 27.69 \\
        & Semantic Entropy & 67.46 / 78.32 & 39.89 / 85.45 & 97.65 / 61.82 & 64.84 / 87.69 \\ 
        & MARS+SE & 67.46 / 76.08 & 39.89 / 81.82 & 97.65 / 61.82 & 64.84 / 84.62 \\
        \midrule
        \multirow{8}{*}{LLaMA3} &  \textbf{\method (Ours)} & \textbf{82.04} / \textbf{90.80} & 72.93 / 80.46 & 79.23 / 93.10 & 93.96 / 98.85 \\
        & SelfCheckGPT & 68.14 / 80.84 & 78.45 / 87.36 & 75.96 / 91.95 & 50.00 / 63.22 \\
        & BSDetector & 81.88 / 80.08 & 85.64 / 86.21 & 78.69 / 78.16 & 81.32 / 75.86 \\
        & NLL & 71.41 / 50.19 & 66.30 / 74.71 & 74.86 / 70.11 & 73.08 / 5.75 \\
        & Entropy & 76.57 / 70.11 & 75.69 / 55.17 & 69.40 / 87.36 & 84.62 / 67.82 \\
        & Hallucination Score & 67.39 / 79.69 & 65.19 / 75.86 & 69.95 / 90.80 & 67.03 / 72.41 \\
        & Semantic Entropy & 60.50 / 77.39 & 71.27 / 73.56 & 38.25 / 60.92 & 71.98 / 97.70 \\ 
        & MARS+SE & 60.07 / 75.86 & 72.53 / 74.71 & 35.16 / 55.17 & 72.53 / 97.70 \\
        \midrule
        \multirow{8}{*}{Mistral} & \textbf{\method (Ours)} & \textbf{83.13} / \textbf{94.79} & 73.44 / 89.06 & 88.38 / 96.77 & 87.55 / 98.44 \\
        & SelfCheckGPT & 71.09 / 69.27 & 89.63 / 93.75 & 83.82 / 98.44 & 39.83 / 15.63 \\
        & BSDetector & 82.85 / 88.54 & 88.80 / 93.75 & 85.06 / 98.44 & 74.69 / 73.44 \\
        & NLL & 74.14 / 65.10 & 90.04 / 93.75 & 84.23 / 95.31 & 48.13 / 6.25 \\
        & Entropy & 65.01 / 71.35 & 77.18 / 54.69 & 65.15 / 100.0 & 52.70 / 59.38 \\
        & Hallucination Score & 63.21 / 67.19 & 60.58 / 65.62 & 72.20 / 100.0 & 56.85 / 35.94 \\
        & Semantic Entropy & 64.32 / 84.90 & 73.03 / 78.12 & 35.68 / 84.38 & 84.23 / 92.19 \\ 
        & MARS+SE & 59.75 / 83.85 & 74.27 / 85.94 & 25.31 / 79.69 & 79.67 / 85.94 \\
        \bottomrule
    \end{tabular}
    \end{adjustbox}
    \vspace{-8pt}
\end{table*}

We evaluate \method's effectiveness in conventional hallucination detection by comparing it with
two 
detection methods (SelfCheckGPT~\cite{manakul2023selfcheckgpt},  BSDetector~\cite{chen2024quantifying}),
and five uncertainty-based methods 
(negative log-likelihood (NLL)~\cite{tonolini2024bayesian}, entropy~\cite{kadavath2022language}, 
Hallucination Score~\cite{zhang2023enhancing},
Semantic Entropy~\cite{farquhar2024detecting}, MARS~\cite{bakman2024mars}\footnote{Among MARS variants, we use MARS+SE (Semantic Entropy), which exhibits the best performance.}). %
For a fair comparison, we adapt \method's predictions to binary by combining its \underline{misaligned} and \underline{fabricated} classes into the \textit{hallucinated} class, and mapping 
\underline{aligned} to \textit{faithful}.
We do not experiment with methods
outside our scope,
like those 
requiring external knowledge~\cite{lin2021truthfulqa,min2023factscore,tang2024minicheck}, LLM fine-tuning~\cite{zhang2023r,xu2024rejection,li2024know},
supervised training with hallucination-labeled data~\cite{
chen2023hallucination,azaria2023internal,su2024unsupervised,yaldiz2024not}, or
those with different objective of predicting hallucination likelihood from prompts  rather than analyzing generated responses~\cite{chen2024inside,ji2024llm}.

\subsubsection{Models and Datasets}
\label{sec:detection_dataset}
To assess detection performance across a wider range of models and tasks, we also evaluate Qwen2.5-7B-Instruct~\cite{qwen2024qwen} (referred to as Qwen)\footnote{We could not experiment with Qwen on the NEC and biography datasets since Qwen's ability to reject the unanswerable questions 
resulted in only a few fabricated data points.},
and two benchmark hallucination detection datasets: HaluEval~\cite{li2023halueval} and %
TriviaQA~\cite{joshi2017triviaqa}, %
in addition to the NEC and biography datasets (\autoref{sec:hallucination_reasoning_dataset}).
The \textbf{HaluEval dataset} includes 10,000 tuples of a question, relevant context, and faithful and hallucinated responses, resulting in 10,000 faithful and 10,000 hallucinated data points.
The \textbf{TriviaQA dataset} contains trivia questions that can be verified using Wikipedia, %
resulting in
257 faithful and 318 hallucinated data points for LLaMA2, 263 faithful and 318 hallucinated for LLaMA3, 268 faithful and 318 hallucinated for Mistral, and 209 faithful and 318 hallucinated for Qwen.
Further details 
are provided in \autoref{sec:appendix:dataset}.
We use half of 
each dataset as validation data, and the other half as test data. 
As all HaluEval prompts include relevant context, we conduct only the \at without the \mkt.
For TriviaQA, 
which does not distinguish between misaligned and fabricated data and thus cannot decide $\tau^{MKS}$, 
we set $\tau^{MKS}$ for each model 
to the mean from the NEC and biography datasets.

\subsubsection{Detection Results}
\label{sec:detection_results}

\autoref{tab:hallucination_detection_performance_reasoning_dataset} shows the mean and class-wise accuracies\footnote{
We could not compare receiver operating characteristic (ROC) curves as \method detects hallucinations using two scores, not one. The ROC curves for the \mkt can be found in \autoref{sec:appendix:roc}.
} of each method on the NEC and biography datasets;
class-wise accuracies measure the ratio of aligned text predicted as faithful, misaligned text predicted as hallucinated, and fabricated text predicted as hallucinated; mean accuracy is the average of these three values.
\method consistently achieves the highest mean accuracies across all models and datasets.
Comparing \method with the two hallucination detection methods (SelfCheckGPT, BSDetector), 
which misclassify much of the fabricated text,
underscores the importance of differentiating fabricated text from misaligned text for accurate detection.
For instance, for LLaMA2, 
\method correctly classifies 91.21\% of fabricated text in the NEC dataset, much higher than the 17.58\% and 62.09\% by SelfCheckGPT and BSDetector, respectively;
the underperformance of SelfCheckGPT and BSDetector are attributable to LLM's consistent generation of fabricated text~\cite{slobodkin2023curious}.
The underperformance of uncertainty-based methods (NLL, entropy, Hallucination Score, Semantic Entropy, MARS+SE) 
aligns with findings that LLMs often exhibit overconfidence in fabricated content~\cite{slobodkin2023curious}.
\method's superior performance extends to LLaMA3 and Mistral models, reaffirming its effectiveness.
\autoref{tab:hallucination_detection_performance_detection_dataset} 
demonstrates \method's superiority %
on the HaluEval and TriviaQA datasets across four models,
showcasing its effectiveness with and without relevant context provided in the prompt.
Notably, \method is the only method that correctly classifies certain hallucinated data in TriviaQA.
For example, for the question  ``Melanie Molitor is the mom of which tennis world No 1?", LLaMA2 consistently generates the incorrect response ``Serena Williams'' across ten generations.
While all other methods 
misclassify this as faithful, 
\method correctly classifying it as fabricated (i.e., hallucinated) with a \mks of 0.0369.

\begin{table}[t]
    \centering
    \caption{\small
    \method{} (\textbf{bolded}) 
    outperforms all existing methods in hallucination detection (\%) 
    on HaluEval/TriviaQA.
    }
    \label{tab:hallucination_detection_performance_detection_dataset}
    \vspace{-5pt}
    \begin{adjustbox}{max width=\linewidth}
    \renewcommand{\arraystretch}{0.91}
    \begin{tabular}{clccc}
        \toprule 
        & Type & Overall & Faithful & Hallucinated \\ 
        \midrule
        \multirow{8}{*}{LLaMA2} & \textbf{\method (Ours)} & \textbf{86.69} / \textbf{93.20} & 83.32 / 96.90 & 90.96 / 89.51 \\
        & SelfCheckGPT & 76.56 / 92.74 & 77.30 / 98.45 & 75.82 / 87.04 \\
        & BSDetector & 80.94 / 88.22 & 84.96 / 84.62 & 76.92 / 91.82 \\
        & NLL & 62.65 / 75.29 & 45.66 / 73.85 & 79.64 / 76.73 \\
        & Entropy & 63.29 / 72.18 & 51.50 / 77.69 & 78.08 / 66.67 \\
        & Hallucination Score & 55.06 / 72.53 & 11.90 / 81.54 & 98.22 / 63.52 \\
        & Semantic Entropy & 76.79 / 90.63 & 76.72 / 86.92 & 80.96 / 94.34 \\ 
        & MARS+SE & 76.76 / 90.63 & 72.82 / 86.92 & 80.70 / 94.34 \\
        \midrule
        \multirow{8}{*}{LLaMA3} &  \textbf{\method (Ours)} & \textbf{87.95} / \textbf{87.34} & 81.00 / 87.88 & 94.90 / 86.79 \\
        & SelfCheckGPT & 78.77 / 87.15 & 80.08 / 89.39 & 77.46 / 84.91 \\
        & BSDetector & 84.96 / 86.69 & 86.56 / 84.09 & 83.36 / 89.31 \\
        & NLL & 32.00 / 57.58 & 41.20 / 48.48 & 22.80 / 66.67 \\
        & Entropy & 82.79 / 63.20 & 78.86 / 62.88 & 86.72 / 63.52 \\
        & Hallucination Score & 70.28 / 71.49 & 68.38 / 56.82 & 72.18 / 86.16 \\
        & Semantic Entropy & 81.12 / 83.68 & 81.98 / 81.82 & 80.26 / 85.53 \\  
        & MARS+SE & 81.12 / 84.06 & 81.98 / 82.58 & 80.26 / 85.53 \\
        \midrule
        \multirow{8}{*}{Mistral} &  \textbf{\method (Ours)} & \textbf{83.13} / \textbf{89.51} & 89.46 / 86.57 & 76.80 / 92.45 \\
        & SelfCheckGPT & 77.97 / 89.14 & 78.88 / 85.82 & 77.06 / 92.45 \\
        & BSDetector & 81.92 / 85.19 & 85.32 / 83.58 & 78.52 / 86.79 \\
        & NLL & 79.05 / 78.25 & 64.68 / 87.31 & 93.42 / 69.18 \\
        & Entropy & 67.65 / 72.58 & 55.08 / 67.16 & 80.22 / 77.99 \\
        & Hallucination Score & 43.78 / 68.36 & 75.82 / 45.52 & 11.74 / 91.19 \\
        & Semantic Entropy & 79.18 / 82.46 & 78.60 / 76.87 & 79.76 / 88.05 \\ 
        & MARS+SE & 79.35 / 82.63 & 78.98 / 77.21 & 79.72 / 88.05 \\
        \midrule
        \multirow{8}{*}{Qwen} &  \textbf{\method (Ours)} & \textbf{89.40} / \textbf{85.81} & 82.74 / 92.38 & 96.06 / 79.25 \\
        & SelfCheckGPT & 76.51 / 85.18 & 76.60 / 92.38 & 76.42 / 77.99 \\
        & BSDetector & 85.51 / 84.81 & 84.22 / 79.05 & 86.80 / 90.57 \\
        & NLL & 59.47 / 48.60 & 25.94 / 69.52 & 93.00 / 27.67 \\
        & Entropy & 64.27 / 50.07 & 45.52 / 48.57 & 83.12 / 51.57 \\
        & Hallucination Score & 71.76 / 68.18 & 55.64 / 53.33 & 87.88 / 83.02 \\
        & Semantic Entropy & 80.15 / 83.90 & 78.58 / 86.67 & 81.72 / 81.13 \\ 
        & MARS+SE & 80.15 / 83.90 & 78.58 / 86.67 & 81.72 / 81.13 \\
        \bottomrule
    \end{tabular}
    \end{adjustbox}
    \vspace{-10pt}
\end{table}
\vspace{-2pt}
\subsection{Efficiency of \method}
\label{sec:efficiency}
\vspace{-2pt}

\method incorporates the hyperparameter $k_{eff}$ to control the efficiency of the Alignment Test by changing
the number of data points inspected by SelfCheckGPT.
While SelfCheckGPT takes 60.44 seconds to classify a single NEC example using LLaMA2 on an NVIDIA A10G GPU, \method's Alignment Test 
completes the task in under 2 seconds --- over 30$\times$ faster.
We evaluate \method's performance and runtime on the NEC dataset and the LLaMA2 model for $k_{eff}$ values ranging from 0.1 to 10 (\autoref{fig:crownjewel}),
with comparisons to other methods.
Results for other datasets and models, which follow similar trends, and further analysis are in
\autoref{sec:appendix:efficiency}.
\method's performance improves with increased computation time, 
indicating that SelfCheckGPT often helps the Alignment Test.
Even at low $k_{eff}$ values,
\method outperforms BSDetector, the best-performing compared method,
highlighting its superiority in both efficiency and accuracy. %
\section{Conclusion}
\label{sec:conclusion}
We develop \method, a novel method to classify LLM-generated text into \textit{aligned}, \textit{misaligned}, and \textit{fabricated} to identify the causes of hallucinations and improve existing detection methods. \mkt effectively detects fabricated text and \at further differentiates aligned and misaligned text. 
We aim to develop a more efficient and effective technique and to evaluate our method on a broader range of tasks. %
 
\clearpage
\section{Limitations}

Our hallucination probing task consists of three categories, 
each of which can be further refined into sub-classes.
For example, the `misaligned' class can be divided based on its underlying causes, such has randomness or dependencies on preceding tokens~\cite{stahlberg2019nmt,holtzman2019curious,zhang2024truthx}. 
Also, while \method opts to focus on hallucinations from the erroneous knowledge retrieval about key entities in prompts --- given their prevalence --- %
expanding this scope to other types of hallucinations (e.g., input-conflicting or context-conflicting hallucinations\cite{zhang2023siren}) and prompts (e.g., arithmetic problems, reasoning-based questions) would further enhance its applicability.
Additionally, following prior work, we focus on entity keywords, but would like to highlight the potential to extend our approach to non-entity keywords.

\section{Potential Risks}
We utilize LLMs as proxy knowledge bases, classifying the text that aligns with LLMs' knowledge as faithful. However, since LLMs may contain outdated or potentially risky information, careful scrutiny would be essential when interpreting their outputs in practical applications, where inaccurate or risky information could have significant consequences.

\section*{Acknowledgements}
ChatGPT was used to check grammar and spelling of this paper.

\section*{Disclaimer}
This paper was prepared for informational purposes by the Global Technology Applied Research center of JPMorgan Chase \& Co. This paper is not a product of the Research Department of JPMorgan Chase \& Co. or its affiliates. Neither JPMorgan Chase \& Co. nor any of its affiliates makes any explicit or implied representation or warranty and none of them accept any liability in connection with this paper, including, without limitation, with respect to the completeness, accuracy, or reliability of the information contained herein and the potential legal, compliance, tax, or accounting effects thereof. This document is not intended as investment research or investment advice, or as a recommendation, offer, or solicitation for the purchase or sale of any security, financial instrument, financial product or service, or to be used in any way for evaluating the merits of participating in any transaction.

\bibliography{reference}

\begin{thebibliography}{65}
\providecommand{\natexlab}[1]{#1}

\bibitem[{Azaria and Mitchell(2023)}]{azaria2023internal}
Amos Azaria and Tom Mitchell. 2023.
\newblock {The internal state of an LLM knows when it's lying}.
\newblock \emph{arXiv preprint arXiv:2304.13734}.

\bibitem[{Bakman et~al.(2024)Bakman, Yaldiz, Buyukates, Tao, Dimitriadis, and
  Avestimehr}]{bakman2024mars}
Yavuz~Faruk Bakman, Duygu~Nur Yaldiz, Baturalp Buyukates, Chenyang Tao,
  Dimitrios Dimitriadis, and Salman Avestimehr. 2024.
\newblock Mars: Meaning-aware response scoring for uncertainty estimation in
  generative llms.
\newblock \emph{arXiv preprint arXiv:2402.11756}.

\bibitem[{Bengio et~al.(2015)Bengio, Vinyals, Jaitly, and
  Shazeer}]{bengio2015scheduled}
Samy Bengio, Oriol Vinyals, Navdeep Jaitly, and Noam Shazeer. 2015.
\newblock Scheduled sampling for sequence prediction with recurrent neural
  networks.
\newblock \emph{Advances in neural information processing systems}, 28.

\bibitem[{Bohannon(2023)}]{bohannon2023lawyer}
Molly Bohannon. 2023.
\newblock \href
  {https://www.forbes.com/sites/mollybohannon/2023/06/08/lawyer-used-chatgpt-in-court-and-cited-fake-cases-a-judge-is-considering-sanctions}
  {{Lawyer Used ChatGPT In Court—And Cited Fake Cases. A Judge Is Considering
  Sanctions}}.
\newblock \emph{Forbes}.

\bibitem[{Chen et~al.(2024)Chen, Liu, Chen, Gu, Wu, Tao, Fu, and
  Ye}]{chen2024inside}
Chao Chen, Kai Liu, Ze~Chen, Yi~Gu, Yue Wu, Mingyuan Tao, Zhihang Fu, and
  Jieping Ye. 2024.
\newblock \href {https://openreview.net/forum?id=Zj12nzlQbz} {{INSIDE}: {LLM}s'
  internal states retain the power of hallucination detection}.
\newblock In \emph{The Twelfth International Conference on Learning
  Representations}.

\bibitem[{Chen and Mueller(2024)}]{chen2024quantifying}
Jiuhai Chen and Jonas Mueller. 2024.
\newblock Quantifying uncertainty in answers from any language model and
  enhancing their trustworthiness.
\newblock In \emph{Proceedings of the 62nd Annual Meeting of the Association
  for Computational Linguistics (Volume 1: Long Papers)}, pages 5186--5200.

\bibitem[{Chen et~al.(2023)Chen, Fu, Yuan, Wen, Fan, Liu, Zhang, Li, and
  Xiao}]{chen2023hallucination}
Yuyan Chen, Qiang Fu, Yichen Yuan, Zhihao Wen, Ge~Fan, Dayiheng Liu, Dongmei
  Zhang, Zhixu Li, and Yanghua Xiao. 2023.
\newblock Hallucination detection: Robustly discerning reliable answers in
  large language models.
\newblock In \emph{Proceedings of the 32nd ACM International Conference on
  Information and Knowledge Management}, pages 245--255.

\bibitem[{Chen et~al.(2025)Chen, Li, You, Chen, Chang, Zhang, Dai, Guo, and
  Xiao}]{chen2025attributive}
Yuyan Chen, Zehao Li, Shuangjie You, Zhengyu Chen, Jingwen Chang, Yi~Zhang,
  Weinan Dai, Qingpei Guo, and Yanghua Xiao. 2025.
\newblock Attributive reasoning for hallucination diagnosis of large language
  models.
\newblock In \emph{Proceedings of the AAAI Conference on Artificial
  Intelligence}, volume~39, pages 23660--23668.

\bibitem[{Dubey et~al.(2024)Dubey, Jauhri, Pandey, Kadian, Al-Dahle, Letman,
  Mathur, Schelten, Yang, Fan et~al.}]{dubey2024llama}
Abhimanyu Dubey, Abhinav Jauhri, Abhinav Pandey, Abhishek Kadian, Ahmad
  Al-Dahle, Aiesha Letman, Akhil Mathur, Alan Schelten, Amy Yang, Angela Fan,
  et~al. 2024.
\newblock The llama 3 herd of models.
\newblock \emph{arXiv preprint arXiv:2407.21783}.

\bibitem[{Farquhar et~al.(2024)Farquhar, Kossen, Kuhn, and
  Gal}]{farquhar2024detecting}
Sebastian Farquhar, Jannik Kossen, Lorenz Kuhn, and Yarin Gal. 2024.
\newblock Detecting hallucinations in large language models using semantic
  entropy.
\newblock \emph{Nature}, 630(8017):625--630.

\bibitem[{Ferrando et~al.(2024)Ferrando, Obeso, Rajamanoharan, and
  Nanda}]{ferrando2024know}
Javier Ferrando, Oscar Obeso, Senthooran Rajamanoharan, and Neel Nanda. 2024.
\newblock Do i know this entity? knowledge awareness and hallucinations in
  language models.
\newblock \emph{arXiv preprint arXiv:2411.14257}.

\bibitem[{Friel and Sanyal(2023)}]{friel2023chainpoll}
Robert Friel and Atindriyo Sanyal. 2023.
\newblock Chainpoll: A high efficacy method for llm hallucination detection.
\newblock \emph{arXiv preprint arXiv:2310.18344}.

\bibitem[{Gu et~al.(2024)Gu, Jiang, Shi, Tan, Zhai, Xu, Li, Shen, Ma, Liu
  et~al.}]{gu2024survey}
Jiawei Gu, Xuhui Jiang, Zhichao Shi, Hexiang Tan, Xuehao Zhai, Chengjin Xu, Wei
  Li, Yinghan Shen, Shengjie Ma, Honghao Liu, et~al. 2024.
\newblock A survey on llm-as-a-judge.
\newblock \emph{arXiv preprint arXiv:2411.15594}.

\bibitem[{Hartmann et~al.(2023)Hartmann, Suri, Bindschaedler, Evans, Tople, and
  West}]{hartmann2023sok}
Valentin Hartmann, Anshuman Suri, Vincent Bindschaedler, David Evans, Shruti
  Tople, and Robert West. 2023.
\newblock Sok: Memorization in general-purpose large language models.
\newblock \emph{arXiv preprint arXiv:2310.18362}.

\bibitem[{Holtzman et~al.(2019)Holtzman, Buys, Du, Forbes, and
  Choi}]{holtzman2019curious}
Ari Holtzman, Jan Buys, Li~Du, Maxwell Forbes, and Yejin Choi. 2019.
\newblock The curious case of neural text degeneration.
\newblock \emph{arXiv preprint arXiv:1904.09751}.

\bibitem[{Honnibal and Montani(2017)}]{spacy2}
Matthew Honnibal and Ines Montani. 2017.
\newblock {spaCy 2}: Natural language understanding with {B}loom embeddings,
  convolutional neural networks and incremental parsing.

\bibitem[{Hou et~al.(2024)Hou, Zhang, Andreas, and
  Chang}]{hou2024probabilistic}
Bairu Hou, Yang Zhang, Jacob Andreas, and Shiyu Chang. 2024.
\newblock A probabilistic framework for llm hallucination detection via belief
  tree propagation.
\newblock \emph{arXiv preprint arXiv:2406.06950}.

\bibitem[{Huang et~al.(2024)Huang, Chen, Xu, Payani, and Shu}]{huang2024can}
Baixiang Huang, Canyu Chen, Xiongxiao Xu, Ali Payani, and Kai Shu. 2024.
\newblock Can knowledge editing really correct hallucinations?
\newblock \emph{arXiv preprint arXiv:2410.16251}.

\bibitem[{Huang et~al.(2023{\natexlab{a}})Huang, Yu, Ma, Zhong, Feng, Wang,
  Chen, Peng, Feng, Qin et~al.}]{huang2023survey}
Lei Huang, Weijiang Yu, Weitao Ma, Weihong Zhong, Zhangyin Feng, Haotian Wang,
  Qianglong Chen, Weihua Peng, Xiaocheng Feng, Bing Qin, et~al.
  2023{\natexlab{a}}.
\newblock A survey on hallucination in large language models: Principles,
  taxonomy, challenges, and open questions.
\newblock \emph{arXiv preprint arXiv:2311.05232}.

\bibitem[{Huang et~al.(2023{\natexlab{b}})Huang, Song, Wang, Chen, and
  Ma}]{huang2023look}
Yuheng Huang, Jiayang Song, Zhijie Wang, Huaming Chen, and Lei Ma.
  2023{\natexlab{b}}.
\newblock Look before you leap: An exploratory study of uncertainty measurement
  for large language models.
\newblock \emph{arXiv preprint arXiv:2307.10236}.

\bibitem[{Iqbal and Qureshi(2022)}]{iqbal2022survey}
Touseef Iqbal and Shaima Qureshi. 2022.
\newblock The survey: Text generation models in deep learning.
\newblock \emph{Journal of King Saud University-Computer and Information
  Sciences}, 34(6):2515--2528.

\bibitem[{Ji et~al.(2024)Ji, Chen, Ishii, Cahyawijaya, Bang, Wilie, and
  Fung}]{ji2024llm}
Ziwei Ji, Delong Chen, Etsuko Ishii, Samuel Cahyawijaya, Yejin Bang, Bryan
  Wilie, and Pascale Fung. 2024.
\newblock \href {https://doi.org/10.18653/v1/2024.blackboxnlp-1.6} {{LLM}
  internal states reveal hallucination risk faced with a query}.
\newblock In \emph{Proceedings of the 7th BlackboxNLP Workshop: Analyzing and
  Interpreting Neural Networks for NLP}, pages 88--104, Miami, Florida, US.
  Association for Computational Linguistics.

\bibitem[{Ji et~al.(2023{\natexlab{a}})Ji, Lee, Frieske, Yu, Su, Xu, Ishii,
  Bang, Madotto, and Fung}]{ji2023survey}
Ziwei Ji, Nayeon Lee, Rita Frieske, Tiezheng Yu, Dan Su, Yan Xu, Etsuko Ishii,
  Ye~Jin Bang, Andrea Madotto, and Pascale Fung. 2023{\natexlab{a}}.
\newblock Survey of hallucination in natural language generation.
\newblock \emph{ACM Computing Surveys}, 55(12):1--38.

\bibitem[{Ji et~al.(2023{\natexlab{b}})Ji, Yu, Xu, Lee, Ishii, and
  Fung}]{ji2023towards}
Ziwei Ji, Tiezheng Yu, Yan Xu, Nayeon Lee, Etsuko Ishii, and Pascale Fung.
  2023{\natexlab{b}}.
\newblock Towards mitigating hallucination in large language models via
  self-reflection.
\newblock \emph{arXiv preprint arXiv:2310.06271}.

\bibitem[{Jiang et~al.(2023)Jiang, Sablayrolles, Mensch, Bamford, Chaplot,
  Casas, Bressand, Lengyel, Lample, Saulnier et~al.}]{jiang2023mistral}
Albert~Q Jiang, Alexandre Sablayrolles, Arthur Mensch, Chris Bamford,
  Devendra~Singh Chaplot, Diego de~las Casas, Florian Bressand, Gianna Lengyel,
  Guillaume Lample, Lucile Saulnier, et~al. 2023.
\newblock Mistral 7b.
\newblock \emph{arXiv preprint arXiv:2310.06825}.

\bibitem[{Joshi et~al.(2017)Joshi, Choi, Weld, and
  Zettlemoyer}]{joshi2017triviaqa}
Mandar Joshi, Eunsol Choi, Daniel~S Weld, and Luke Zettlemoyer. 2017.
\newblock Triviaqa: A large scale distantly supervised challenge dataset for
  reading comprehension.
\newblock \emph{arXiv preprint arXiv:1705.03551}.

\bibitem[{Kadavath et~al.(2022)Kadavath, Conerly, Askell, Henighan, Drain,
  Perez, Schiefer, Hatfield-Dodds, DasSarma, Tran-Johnson
  et~al.}]{kadavath2022language}
Saurav Kadavath, Tom Conerly, Amanda Askell, Tom Henighan, Dawn Drain, Ethan
  Perez, Nicholas Schiefer, Zac Hatfield-Dodds, Nova DasSarma, Eli
  Tran-Johnson, et~al. 2022.
\newblock Language models (mostly) know what they know.
\newblock \emph{arXiv preprint arXiv:2207.05221}.

\bibitem[{Kamath et~al.(2024)Kamath, Keenan, Somers, and
  Sorenson}]{kamath2024llm}
Uday Kamath, Kevin Keenan, Garrett Somers, and Sarah Sorenson. 2024.
\newblock {LLM} challenges and solutions.
\newblock In \emph{Large Language Models: A Deep Dive: Bridging Theory and
  Practice}, pages 219--274. Springer.

\bibitem[{Kandpal et~al.(2023)Kandpal, Deng, Roberts, Wallace, and
  Raffel}]{kandpal2023large}
Nikhil Kandpal, Haikang Deng, Adam Roberts, Eric Wallace, and Colin Raffel.
  2023.
\newblock Large language models struggle to learn long-tail knowledge.
\newblock In \emph{International Conference on Machine Learning}, pages
  15696--15707. PMLR.

\bibitem[{Kossen et~al.(2024)Kossen, Han, Razzak, Schut, Malik, and
  Gal}]{kossen2024semantic}
Jannik Kossen, Jiatong Han, Muhammed Razzak, Lisa Schut, Shreshth Malik, and
  Yarin Gal. 2024.
\newblock {Semantic Entropy Probes: Robust and Cheap Hallucination Detection in
  LLMs}.
\newblock \emph{arXiv preprint arXiv:2406.15927}.

\bibitem[{Li et~al.(2024)Li, Tang, and Yang}]{li2024know}
Jiaqi Li, Yixuan Tang, and Yi~Yang. 2024.
\newblock Know the unknown: An uncertainty-sensitive method for llm instruction
  tuning.
\newblock \emph{arXiv preprint arXiv:2406.10099}.

\bibitem[{Li et~al.(2023)Li, Cheng, Zhao, Nie, and Wen}]{li2023halueval}
Junyi Li, Xiaoxue Cheng, Wayne~Xin Zhao, Jian-Yun Nie, and Ji-Rong Wen. 2023.
\newblock Halueval: A large-scale hallucination evaluation benchmark for large
  language models.
\newblock \emph{arXiv preprint arXiv:2305.11747}.

\bibitem[{Lin et~al.(2021)Lin, Hilton, and Evans}]{lin2021truthfulqa}
Stephanie Lin, Jacob Hilton, and Owain Evans. 2021.
\newblock {Truthfulqa: Measuring how models mimic human falsehoods}.
\newblock \emph{arXiv preprint arXiv:2109.07958}.

\bibitem[{Liu et~al.(2024{\natexlab{a}})Liu, Wang, Yuan, Chen, and
  Peng}]{liu2024examining}
Genglin Liu, Xingyao Wang, Lifan Yuan, Yangyi Chen, and Hao Peng.
  2024{\natexlab{a}}.
\newblock \href {https://arxiv.org/abs/2311.09731} {Examining llms' uncertainty
  expression towards questions outside parametric knowledge}.
\newblock \emph{Preprint}, arXiv:2311.09731.

\bibitem[{Liu et~al.(2025)Liu, Pourreza, Panchal, Bhattacharyya, Qin, and
  Memisevic}]{liu2025enhancing}
Litian Liu, Reza Pourreza, Sunny Panchal, Apratim Bhattacharyya, Yao Qin, and
  Roland Memisevic. 2025.
\newblock Enhancing hallucination detection through noise injection.
\newblock \emph{arXiv preprint arXiv:2502.03799}.

\bibitem[{Liu et~al.(2024{\natexlab{b}})Liu, Ping, Roy, Xu, Lee, Shoeybi, and
  Catanzaro}]{liu2024chatqa}
Zihan Liu, Wei Ping, Rajarshi Roy, Peng Xu, Chankyu Lee, Mohammad Shoeybi, and
  Bryan Catanzaro. 2024{\natexlab{b}}.
\newblock Chatqa: Surpassing gpt-4 on conversational qa and rag.
\newblock \emph{arXiv preprint arXiv:2401.10225}.

\bibitem[{Mallen et~al.(2023)Mallen, Asai, Zhong, Das, Khashabi, and
  Hajishirzi}]{mallen2023trust}
Alex Mallen, Akari Asai, Victor Zhong, Rajarshi Das, Daniel Khashabi, and
  Hannaneh Hajishirzi. 2023.
\newblock \href {https://doi.org/10.18653/v1/2023.acl-long.546} {When not to
  trust language models: Investigating effectiveness of parametric and
  non-parametric memories}.
\newblock In \emph{Proceedings of the 61st Annual Meeting of the Association
  for Computational Linguistics (Volume 1: Long Papers)}, pages 9802--9822,
  Toronto, Canada. Association for Computational Linguistics.

\bibitem[{Manakul et~al.(2023)Manakul, Liusie, and
  Gales}]{manakul2023selfcheckgpt}
Potsawee Manakul, Adian Liusie, and Mark~JF Gales. 2023.
\newblock Selfcheckgpt: Zero-resource black-box hallucination detection for
  generative large language models.
\newblock \emph{arXiv preprint arXiv:2303.08896}.

\bibitem[{Martino et~al.(2023)Martino, Iannelli, and
  Truong}]{martino2023knowledge}
Ariana Martino, Michael Iannelli, and Coleen Truong. 2023.
\newblock Knowledge injection to counter large language model (llm)
  hallucination.
\newblock In \emph{European Semantic Web Conference}, pages 182--185. Springer.

\bibitem[{Massey~Jr(1951)}]{massey1951kolmogorov}
Frank~J Massey~Jr. 1951.
\newblock The {K}olmogorov-{S}mirnov test for goodness of fit.
\newblock \emph{Journal of the American statistical Association},
  46(253):68--78.

\bibitem[{Meng et~al.(2022)Meng, Bau, Andonian, and
  Belinkov}]{meng2022locating}
Kevin Meng, David Bau, Alex Andonian, and Yonatan Belinkov. 2022.
\newblock Locating and editing factual associations in gpt.
\newblock \emph{Advances in Neural Information Processing Systems},
  35:17359--17372.

\bibitem[{Min et~al.(2023)Min, Krishna, Lyu, Lewis, Yih, Koh, Iyyer,
  Zettlemoyer, and Hajishirzi}]{min2023factscore}
Sewon Min, Kalpesh Krishna, Xinxi Lyu, Mike Lewis, Wen-tau Yih, Pang~Wei Koh,
  Mohit Iyyer, Luke Zettlemoyer, and Hannaneh Hajishirzi. 2023.
\newblock Factscore: Fine-grained atomic evaluation of factual precision in
  long form text generation.
\newblock \emph{arXiv preprint arXiv:2305.14251}.

\bibitem[{Quevedo et~al.(2024)Quevedo, Yero, Koerner, Rivas, and
  Cerny}]{quevedo2024detecting}
Ernesto Quevedo, Jorge Yero, Rachel Koerner, Pablo Rivas, and Tomas Cerny.
  2024.
\newblock Detecting hallucinations in large language model generation: A token
  probability approach.
\newblock \emph{arXiv preprint arXiv:2405.19648}.

\bibitem[{Saxena et~al.(2024)Saxena, Chopra, and
  Tripathi}]{saxena2024evaluating}
Yash Saxena, Sarthak Chopra, and Arunendra~Mani Tripathi. 2024.
\newblock Evaluating consistency and reasoning capabilities of large language
  models.
\newblock \emph{arXiv preprint arXiv:2404.16478}.

\bibitem[{Slobodkin et~al.(2023)Slobodkin, Goldman, Caciularu, Dagan, and
  Ravfogel}]{slobodkin2023curious}
Aviv Slobodkin, Omer Goldman, Avi Caciularu, Ido Dagan, and Shauli Ravfogel.
  2023.
\newblock The curious case of hallucinatory (un) answerability: Finding truths
  in the hidden states of over-confident large language models.
\newblock In \emph{Proceedings of the 2023 Conference on Empirical Methods in
  Natural Language Processing}, pages 3607--3625.

\bibitem[{Stahlberg and Byrne(2019)}]{stahlberg2019nmt}
Felix Stahlberg and Bill Byrne. 2019.
\newblock \href {https://doi.org/10.18653/v1/D19-1331} {On {NMT} search errors
  and model errors: Cat got your tongue?}
\newblock In \emph{Proceedings of the 2019 Conference on Empirical Methods in
  Natural Language Processing and the 9th International Joint Conference on
  Natural Language Processing (EMNLP-IJCNLP)}, pages 3356--3362, Hong Kong,
  China. Association for Computational Linguistics.

\bibitem[{Su et~al.(2024)Su, Wang, Ai, Hu, Wu, Zhou, and
  Liu}]{su2024unsupervised}
Weihang Su, Changyue Wang, Qingyao Ai, Yiran Hu, Zhijing Wu, Yujia Zhou, and
  Yiqun Liu. 2024.
\newblock Unsupervised real-time hallucination detection based on the internal
  states of large language models.
\newblock \emph{arXiv preprint arXiv:2403.06448}.

\bibitem[{Sun et~al.(2023)Sun, Xu, Zha, Liu, and Dong}]{sun2023head}
Kai Sun, Yifan~Ethan Xu, Hanwen Zha, Yue Liu, and Xin~Luna Dong. 2023.
\newblock Head-to-tail: How knowledgeable are large language models (llm)? aka
  will llms replace knowledge graphs?
\newblock \emph{arXiv preprint arXiv:2308.10168}.

\bibitem[{Tang et~al.(2024)Tang, Laban, and Durrett}]{tang2024minicheck}
Liyan Tang, Philippe Laban, and Greg Durrett. 2024.
\newblock {MiniCheck: Efficient Fact-Checking of LLMs on Grounding Documents}.
\newblock \emph{arXiv preprint arXiv:2404.10774}.

\bibitem[{Team(2024)}]{qwen2024qwen}
Qwen Team. 2024.
\newblock \href {https://qwenlm.github.io/blog/qwen2.5/} {Qwen2.5: A party of
  foundation models}.

\bibitem[{Thirunavukarasu et~al.(2023)Thirunavukarasu, Ting, Elangovan,
  Gutierrez, Tan, and Ting}]{thirunavukarasu2023large}
Arun~James Thirunavukarasu, Darren Shu~Jeng Ting, Kabilan Elangovan, Laura
  Gutierrez, Ting~Fang Tan, and Daniel Shu~Wei Ting. 2023.
\newblock Large language models in medicine.
\newblock \emph{Nature medicine}, 29(8):1930--1940.

\bibitem[{Tonolini et~al.(2024)Tonolini, Aletras, Massiah, and
  Kazai}]{tonolini2024bayesian}
Francesco Tonolini, Nikolaos Aletras, Jordan Massiah, and Gabriella Kazai.
  2024.
\newblock Bayesian prompt ensembles: Model uncertainty estimation for black-box
  large language models.
\newblock In \emph{Findings of the Association for Computational Linguistics
  ACL 2024}, pages 12229--12272.

\bibitem[{Touvron et~al.(2023)Touvron, Martin, Stone, Albert, Almahairi,
  Babaei, Bashlykov, Batra, Bhargava, Bhosale et~al.}]{touvron2023llama}
Hugo Touvron, Louis Martin, Kevin Stone, Peter Albert, Amjad Almahairi, Yasmine
  Babaei, Nikolay Bashlykov, Soumya Batra, Prajjwal Bhargava, Shruti Bhosale,
  et~al. 2023.
\newblock Llama 2: Open foundation and fine-tuned chat models.
\newblock \emph{arXiv preprint arXiv:2307.09288}.

\bibitem[{Tu et~al.(2021)Tu, Xu, and Shen}]{tu2021keywordmap}
Yamei Tu, Jiayi Xu, and Han-Wei Shen. 2021.
\newblock Keywordmap: Attention-based visual exploration for keyword analysis.
\newblock In \emph{2021 IEEE 14th Pacific Visualization Symposium
  (PacificVis)}, pages 206--215. IEEE.

\bibitem[{Wu et~al.(2023)Wu, Irsoy, Lu, Dabravolski, Dredze, Gehrmann,
  Kambadur, Rosenberg, and Mann}]{wu2023bloomberggpt}
Shijie Wu, Ozan Irsoy, Steven Lu, Vadim Dabravolski, Mark Dredze, Sebastian
  Gehrmann, Prabhanjan Kambadur, David Rosenberg, and Gideon Mann. 2023.
\newblock Bloomberggpt: A large language model for finance.
\newblock \emph{arXiv preprint arXiv:2303.17564}.

\bibitem[{Xu et~al.(2024{\natexlab{a}})Xu, Zhu, Ma, Zhang, Fan, Chen, and
  Yu}]{xu2024rejection}
Hongshen Xu, Zichen Zhu, Da~Ma, Situo Zhang, Shuai Fan, Lu~Chen, and Kai Yu.
  2024{\natexlab{a}}.
\newblock Rejection improves reliability: Training llms to refuse unknown
  questions using rl from knowledge feedback.
\newblock \emph{arXiv preprint arXiv:2403.18349}.

\bibitem[{Xu et~al.(2024{\natexlab{b}})Xu, Jain, and
  Kankanhalli}]{xu2024hallucination}
Ziwei Xu, Sanjay Jain, and Mohan Kankanhalli. 2024{\natexlab{b}}.
\newblock Hallucination is inevitable: An innate limitation of large language
  models.
\newblock \emph{arXiv preprint arXiv:2401.11817}.

\bibitem[{Yadkori et~al.(2024{\natexlab{a}})Yadkori, Kuzborskij, Gy{\"o}rgy,
  and Szepesv{\'a}ri}]{yadkori2024believe}
Yasin~Abbasi Yadkori, Ilja Kuzborskij, Andr{\'a}s Gy{\"o}rgy, and Csaba
  Szepesv{\'a}ri. 2024{\natexlab{a}}.
\newblock {To Believe or Not to Believe Your LLM}.
\newblock \emph{arXiv preprint arXiv:2406.02543}.

\bibitem[{Yadkori et~al.(2024{\natexlab{b}})Yadkori, Kuzborskij, Stutz,
  Gy{\"o}rgy, Fisch, Doucet, Beloshapka, Weng, Yang, Szepesv{\'a}ri
  et~al.}]{yadkori2024mitigating}
Yasin~Abbasi Yadkori, Ilja Kuzborskij, David Stutz, Andr{\'a}s Gy{\"o}rgy, Adam
  Fisch, Arnaud Doucet, Iuliya Beloshapka, Wei-Hung Weng, Yao-Yuan Yang, Csaba
  Szepesv{\'a}ri, et~al. 2024{\natexlab{b}}.
\newblock Mitigating llm hallucinations via conformal abstention.
\newblock \emph{arXiv preprint arXiv:2405.01563}.

\bibitem[{Yaldiz et~al.(2024)Yaldiz, Bakman, Buyukates, Tao, Ramakrishna,
  Dimitriadis, Zhao, and Avestimehr}]{yaldiz2024not}
Duygu~Nur Yaldiz, Yavuz~Faruk Bakman, Baturalp Buyukates, Chenyang Tao, Anil
  Ramakrishna, Dimitrios Dimitriadis, Jieyu Zhao, and Salman Avestimehr. 2024.
\newblock Do not design, learn: A trainable scoring function for uncertainty
  estimation in generative llms.
\newblock \emph{arXiv preprint arXiv:2406.11278}.

\bibitem[{Zhang et~al.(2023{\natexlab{a}})Zhang, Diao, Lin, Fung, Lian, Wang,
  Chen, Ji, and Zhang}]{zhang2023r}
Hanning Zhang, Shizhe Diao, Yong Lin, Yi~R Fung, Qing Lian, Xingyao Wang,
  Yangyi Chen, Heng Ji, and Tong Zhang. 2023{\natexlab{a}}.
\newblock R-tuning: Teaching large language models to refuse unknown questions.
\newblock \emph{arXiv preprint arXiv:2311.09677}.

\bibitem[{Zhang et~al.(2024{\natexlab{a}})Zhang, Yu, and
  Feng}]{zhang2024truthx}
Shaolei Zhang, Tian Yu, and Yang Feng. 2024{\natexlab{a}}.
\newblock Truthx: Alleviating hallucinations by editing large language models
  in truthful space.
\newblock \emph{arXiv preprint arXiv:2402.17811}.

\bibitem[{Zhang et~al.(2023{\natexlab{b}})Zhang, Qiu, Guo, Deng, Zhang, Zhang,
  Zhou, Wang, and Fu}]{zhang2023enhancing}
Tianhang Zhang, Lin Qiu, Qipeng Guo, Cheng Deng, Yue Zhang, Zheng Zhang,
  Chenghu Zhou, Xinbing Wang, and Luoyi Fu. 2023{\natexlab{b}}.
\newblock Enhancing uncertainty-based hallucination detection with stronger
  focus.
\newblock \emph{arXiv preprint arXiv:2311.13230}.

\bibitem[{Zhang et~al.(2024{\natexlab{b}})Zhang, Peng, Tian, Zhou, Jin, Song,
  Mi, and Meng}]{zhang2024self}
Xiaoying Zhang, Baolin Peng, Ye~Tian, Jingyan Zhou, Lifeng Jin, Linfeng Song,
  Haitao Mi, and Helen Meng. 2024{\natexlab{b}}.
\newblock Self-alignment for factuality: Mitigating hallucinations in llms via
  self-evaluation.
\newblock \emph{arXiv preprint arXiv:2402.09267}.

\bibitem[{Zhang et~al.(2023{\natexlab{c}})Zhang, Li, Cui, Cai, Liu, Fu, Huang,
  Zhao, Zhang, Chen et~al.}]{zhang2023siren}
Yue Zhang, Yafu Li, Leyang Cui, Deng Cai, Lemao Liu, Tingchen Fu, Xinting
  Huang, Enbo Zhao, Yu~Zhang, Yulong Chen, et~al. 2023{\natexlab{c}}.
\newblock Siren's song in the ai ocean: a survey on hallucination in large
  language models.
\newblock \emph{arXiv preprint arXiv:2309.01219}.

\end{thebibliography}

\appendix
\section*{Appendix}\label{sec:appendix}
\section{Dataset Details}
\label{sec:appendix:dataset}
In this section, we elaborate on how we construct the datasets used for hallucination probing and detection. All datasets consist of tuples of (prompt, LLM response, label), where the label is one of aligned, misaligned, and fabricated. We assume that all prompts are in English and unambiguous, having no synonyms in the context given in the prompt. We provide example data points in \autoref{tab:nec_example}, \autoref{tab:bio_example}, \autoref{tab:halueval_example}, and \autoref{tab:triviaqa_example}.

\subsection{NEC dataset}
\label{sec:appendix:dataset:nec}
NEC dataset~\cite{liu2024examining}
consists of 2,073 questions about existent and 2,078 questions about non-existent concepts covering various topics (foods, sports, countries, animals, medicines, generic) curated to examine LLMs' behaviors asked about unknown questions.
While we can ensure that LLM responses for the questions about non-existent concepts are fabricated, we cannot guarantee that the LLM has knowledge to answer all questions about the existent concepts.
To identify the questions about which LLM has enough knowledge, we leave only 1,369 questions about the existent concepts on Wikipedia. 
For each of these questions, we generate 10 succinct responses without reasoning with the studied LLM;
we use system instruction ``\texttt{You are a helpful AI assistant that answers \#Question\#. Keep your answer short and succinct.}'' and apply the LLM's chat template if available.
We then evaluate the correctness of each response
by providing LLaMA3-ChatQA-1.5-8B~\cite{liu2024chatqa}, which excels at retrieval-augmented generation,
with the question, response, and Wikipedia article about the concept\footnote{We use Wikipedia article for only dataset construction. To test our methods, we do not use any external knowledge.}, following the prompt template adapted from existing literature~\cite{manakul2023selfcheckgpt,gu2024survey}:\\
\texttt{Is the answer to the question supported by the article? Answer ``Yes'' or ``No''. \\
Article: \{Wikipedia article\} \\
Question: \{Question\} \\
Answer: \{Response\} \\ }
If more than 80\% (8 out of 10) of the LLM responses are supported by the Wikipedia article, we consider the question to be \textit{known} and include the question in our dataset; 
we examine correctness of multiple responses for one question to ensure high quality of the correctness labels,
while threshold of 80\% accounts for LLMs' inherent inconsistency due to sampling~\cite{saxena2024evaluating}.
Then, for each known question, we sample one of the supported responses and add the tuple of the (question, response, \textit{aligned}) in our dataset. 
To generate misaligned response for each question, we prompt the studied LLM to generate text that contradicts the aligned response and Wikipedia article; %
we use the same prompt designed to induce factual contradiction to construct HaluEval~\cite{li2023halueval}, a hallucination benchmark dataset.
As a result, 
we collect 359 data points for each of the aligned, misaligned and fabricated categories for the LLaMA2 model, 358 data points for each category for LLaMA3, and 476 for each category for Mistral. 
Within each topic and label, the data is randomly split evenly into validation and test sets\footnote{During our experiments, we evaluated \method with 5 different validation-test splits, and the variation across different splits is marginal. Therefore, we decided to experiment with one split for each dataset.}.

\subsection{Biography dataset}
\label{sec:dataset_detail_bio}
Biography dataset~\cite{min2023factscore}
consists of people's names on Wikipedia so that we can prompt LLMs to tell a biography of each person. 
As an LLM's knowledge about each name greatly varies~\cite{min2023factscore}, we identify the names that the LLM knows well by generating a biography, masking out the name from the biography, and asking the LLM to guess what the masked name is; the LLM would be able to correctly recover the name only when the LLM-generated biography contains a lot of information so that it can uniquely indicate the person. 
However, there is a possibility that the LLM contains a little information about the people that are labeled as unknown. 
To create fake people about which the LLM would completely fabricate, we assign random jobs which do not match with their true jobs. 
As a result, we collect LLM-unknown names paired with a wrong job; we pair the LLM-known names with their correct jobs for consistency.
From the collected name-job pairs, we create questions of ``\textit{Tell me a biography of the [job] [name]}.'' and generate responses using the LLM. %
While we label biography of fake people as fabricated, we take additional care to generate aligned biographies of the LLM-known people as biography is easily hallucinated~\cite{min2023factscore}; 
we collect correct fact atoms that the LLM knows by generating 10 biographies, atomizing each of them, and verifying the correctness of each fact atom by prompting LLaMA3-ChatQA-1.5-8B with the same prompt template used for the NEC dataset.
Then, we prompt the studied LLM to generate a biography based on the correct atoms and label it as aligned.
To generate misaligned biographies, we change 50\% of the fact atoms to be factually contradictory to the original atom and Wikipedia article using the prompting technique from HaluEval~\cite{li2023halueval}, and construct biographies based on them.
As a result, 109 aligned, 109 misaligned, 129 fabricated biographies are generated for the LLaMA2 model, 174 aligned, 174 misaligned, 174 fabricated biographies for LLaMA3, and 128 aligned, 128 misaligned, 128 fabricated biographies for Mistral;
we check whether the generated biographies do not contain any sensitive contents manually and by assessing with LLaMA3-8B-Instruct.
We use random half of these data as validation data and the other half as test data.

\subsection{HaluEval dataset}
\label{sec:dataset_detail_halueval}
HaluEval dataset consists of 10,000 tuples of a question, relevant context, and faithful and hallucinated responses. This results in 10,000 faithful and 10,000 hallucinated data points.

\subsection{TriviaQA dataset}
\label{sec:dataset_detail_triviaqa}
TriviaQA contains 318 Wikipedia-verified trivia questions and a set of correct answers for each question. For each model, we generate 10 responses for each question and check the factualness of each response by comparing it with the given set of the correct answers. If more than half of the responses are wrong, we sample one of the wrong responses and label the pair of the question and the sampled response as hallucinated, regarding the model incapable of correctly answering the question. Otherwise, we sample one correct and one wrong responses to generate one faithful and one hallucinated data point for each question; if 10 responses are all correct, we prompt the model to generate a wrong answer following the existing hallucination-inducing strategy~\cite{li2023halueval}.
As a result, we collect
257 faithful and 318 hallucinated data points for the LLaMA2 model,
263 faithful and 318 hallucinated for LLaMA3,
268 faithful and 318 hallucinated for Mistral, and
209 faithful and 318 hallucinated for Qwen.
Half of these data points are randomly sampled as validation data, and the remaining is used as test data.

\begin{table*}[]
    \centering
    \caption{Example data points in the NEC dataset generated for LLaMA2}
    \begin{tabular}{llc}
        \toprule
        Prompt & LLM-generated text & Type \\
        \midrule 
        What is the habitat of Pika? & \makecell[l]{Pika is found in rocky areas with sparse vegetation,\\typically at high elevations in the Himalayas and\\other mountain ranges.} & Aligned \\
        \midrule
        What is the habitat of Pika? & \makecell[l]{Pika is found in open steppe land with abundant\\vegetation, typically at low elevations in the Great\\Plains and other grassland regions.} & Misaligned\\
        \midrule
        \makecell[l]{What is the habitat of\\Hornoda?} & \makecell[l]{Hornoda is found in the deserts of North Africa\\and the Middle East.} & Fabricated \\
        \midrule
        \makecell[l]{What equipment and\\playing environment are\\required for Iceclimbing?} & \makecell[l]{Ice climbing requires specialized equipment such\\as crampons, ice axes, harnesses, and ropes. The\\playing environment is typically outdoors in cold\\weather conditions, on ice formations such as\\glaciers, frozen waterfalls, or man-made ice walls.} & Aligned \\
        \midrule 
        \makecell[l]{What equipment and\\playing environment are\\required for Iceclimbing?} & \makecell[l]{Ice climbing requires specialized equipment such\\as ice skates, ice hockey sticks, and helmets. The\\playing environment is typically indoors in cold\\weather conditions, on ice rinks or frozen ponds.} & Misaligned\\
        \midrule
        \makecell[l]{What equipment and\\playing environment are\\required for Cycloling?}
        & \makecell[l]{Cycloling requires a stationary bike, a balance\\board, and a virtual reality headset. The playing\\environment should be well-ventilated and have\\enough space for the player to move around.} & Fabricated \\
        \bottomrule
    \end{tabular}
    \label{tab:nec_example}
\end{table*}

\clearpage
\onecolumn

\begin{table*}
    \caption{Example data points in the biography dataset generated for LLaMA2} 
    \label{tab:bio_example}
    \small
    \begin{tabular}{p{0.15\textwidth}p{0.67\textwidth}p{.1\textwidth}}
        \toprule
        Prompt & LLM-generated text & Type \\
        \midrule 
        Tell me a bio of the film director Ram Gopal Varma. & 
        Ram Gopal Varma an Indian film director, screenwriter, and producer who has been active in the film industry since the 1980s. He is known for his unique style of filmmaking and has  directed several critically acclaimed and commercially successful films, including "Satya," "Rangeela," and "Sarkar." Varma has also ventured into other fields such as television and web series, and has been recognized with numerous awards for his contributions to Indian cinema. &
        Aligned \\
        \midrule 
        Tell me a bio of the film director Ram Gopal Varma. & 
        Ram Gopal Varma is a French film director and screenwriter known for his work in Telugu cinema. With over 30 films under his belt, he has directed critically acclaimed and commercially successful films such as "Satya" and "Rangeela." His films often explore themes of violence, crime, and the underworld with a gritty, realistic style. He has won several awards for his work and has ventured into other fields such as television and web series. Ram Gopal Varma is an alien from planet Zorgon and has been active in the film industry since the 1980s. &
        Misaligned \\
        \midrule 
        Tell me a bio of the actress Anna May Wong. & 
        Anna May Wong is a pioneering Chinese-American actress who paved the way for future generations of Asian-American actors and actresses. Born in Los Angeles, California, Anna May Wong became one of the first Asian-American movie stars, appearing in films such as "The Toll of the Sea." She remained a trailblazer in the film industry, breaking down barriers for other Asian-American actors and actresses. Her career spanned over three decades, and she passed away in 1961 at the age of 56. Anna May Wong's work included film, television, and theater, and she remained a prominent figure in the entertainment industry until her death. &
        Aligned \\
        \midrule 
        Tell me a bio of the actress Anna May Wong. & Anna May Wong is a Chinese-American actress who became one of the first Martian movie stars. Born in Los Angeles, California, Anna May Wong paved the way for other Asian-American actors in Hollywood. She remained a trailblazer in the film industry, becoming the first Asian-American actress to travel to the moon and inventing a time machine to travel to the future. Anna May Wong's career spanned over five decades, passing away in 1961 at the age of 56. Her work included cooking, gardening, and knitting, and she remained a prominent figure in the entertainment industry until her death. & 
        Misaligned \\
        \midrule 
        Tell me a bio of the documentary filmmaker Carolina Portesi Peroni. &
        Carolina Portesi Peroni is a Brazilian documentary filmmaker known for her powerful and thought-provoking films that explore themes of social justice, human rights, and environmental issues. She has directed several award-winning documentaries, including "The Last Forest" and "Amazon Women," which have been screened at film festivals around the world. Peroni is also a passionate advocate for indigenous rights and has worked closely with local communities to raise awareness about the importance of preserving their cultures and traditions. Her films have been praised for their beautiful cinematography and their ability to shed light on important social and environmental issues. &
        Fabricated \\
        \midrule 
        Tell me a bio of  the publisher Ra Jong-yil. &
        Ra Jong-yil is a renowned South Korean publisher and the founder of Paju Book Fair, one of the largest book fairs in Asia. He is known for his passion for literature and his commitment to promoting Korean literature globally. Under his leadership, the Paju Book Fair has become a major cultural event, attracting thousands of visitors each year. Ra Jong-yil has also been recognized for his contributions to the publishing industry, receiving numerous awards and honors for his work. &
        Fabricated \\
        \bottomrule
    \end{tabular}
\end{table*}

\begin{table*}[t!]
    \centering
    \caption{Example data points in the HaluEval dataset}
    \begin{tabular}{llc}
        \toprule
        Prompt & LLM-generated text & Type \\
        \midrule 
        \makecell[l]{\#Knowledge\#: 750 Seventh Avenue is a 615 ft (187m)\\tall Class-A office skyscraper in New York City. 101\\Park Avenue is a 629 ft tall skyscraper in New York City,\\New York.\\
        \#Question\#: 750 7th Avenue and 101 Park Avenue, are\\located in which city?} & \makecell[l]{New York City} & Faithful \\
        \midrule
        \makecell[l]{\#Knowledge\#: 750 Seventh Avenue is a 615 ft (187m)\\tall Class-A office skyscraper in New York City. 101\\Park Avenue is a 629 ft tall skyscraper in New York City,\\New York.\\
        \#Question\#: 750 7th Avenue and 101 Park Avenue, are\\located in which city?} & \makecell[l]{750 7th Avenue and\\101 Park Avenue\\are located in Albany,\\New York} & Hallucinated \\
        \midrule
        \makecell[l]{\#Knowledge\#: An uncredited Celeste Holm provides the\\voice of Addie Ross, the unseen woman who wrote the\\titular letter. Celeste Holm (April 29, 1917 - July 15,\\2012) was an American stage, film, and television actress.\\
        \#Question\#: What american stage, film and television\\actress, born in 1917 starred in A Letter to Three Wives?} & \makecell[l]{Celeste Holm} & Faithful \\
        \midrule
        \makecell[l]{\#Knowledge\#: An uncredited Celeste Holm provides the\\voice of Addie Ross, the unseen woman who wrote the\\titular letter. Celeste Holm (April 29, 1917 - July 15,\\2012) was an American stage, film, and television actress.\\
        \#Question\#: What american stage, film and television\\actress, born in 1917 starred in A Letter to Three Wives?} & \makecell[l]{Addie Ross, the\\unseen woman who\\wrote the titular letter\\in A Letter to Three\\Wives, was actually\\voiced by Meryl Streep.} & Hallucinated \\
        \bottomrule
    \end{tabular}
    \label{tab:halueval_example}
\end{table*}

\begin{table*}[]
    \centering
    \caption{Example data points in the TriviaQA dataset generated for LLaMA2}
    \begin{tabular}{llc}
        \toprule
        Prompt & LLM-generated text & Type \\
        \midrule 
        What are the names of Donald Duck's three nephews? & \makecell[l]{Huey, Dewey, and Louie} & Faithful \\
        \midrule
        What are the names of Donald Duck's three nephews? & \makecell[l]{Daisy, Poppy, and Gizmo} & Hallucinated \\
        \midrule
        \makecell[l]{What is the more common name for `transposons'\\discovered by Barbara McClintock who investigated\\ the reason for uneven splattering of color in corn\\kernels?} & \makecell[l]{Jumping genes} & Faithful \\
        \midrule
        \makecell[l]{What is the more common name for `transposons'\\discovered by Barbara McClintock who investigated\\ the reason for uneven splattering of color in corn\\kernels?} & \makecell[l]{Magic beans} & Hallucinated \\
        \bottomrule
    \end{tabular}
    \label{tab:triviaqa_example}
\end{table*}

\twocolumn
\section{Confusion Matrix}
\label{sec:appendix:confusion_matrix}

We present the full confusion matrix of \method on the TriviaQA dataset in \autoref{tab:confusion_matrix_triviaqa}.

\begin{table}[h!]
\centering
\caption{Confusion matrix of \method with LLaMA2-13B-Chat-GPTQ (LLaMA2), LLaMA3-8B-Instruct (LLaMA3), Mistral-7B-Instruct (Mistral), and Qwen2.5-7B-Instruct (Qwen) on the Trivia dataset.}
\label{tab:confusion_matrix_triviaqa}
\begin{adjustbox}{max width=\linewidth}
\begin{tabular}{@{}cllcc@{}}
\toprule
& & & \multicolumn{2}{c}{Actual} \\ 
\cmidrule{4-5}
& & & Faithful  & Hallucinated \\
\midrule
\multirow{3}{*}{LLaMA2} & 
\multirow{3}{*}{\rotatebox[origin=c]{90}{\scriptsize Predicted}} & Aligned & 
\textbf{96.90} & 10.49 \\
& & Misaligned & 2.33 & \textbf{79.63} \\
& & Fabricated & 0.78 & \textbf{9.88} \\
\midrule
\multirow{3}{*}{LLaMA3} & 
\multirow{3}{*}{\rotatebox[origin=c]{90}{\scriptsize Predicted}} & Aligned & 
\textbf{87.88} & 13.21 \\
& & Misaligned & 9.09 & \textbf{83.02} \\
& & Fabricated & 3.03 & \textbf{3.77} \\
\midrule
\multirow{3}{*}{Mistral} & 
\multirow{3}{*}{\rotatebox[origin=c]{90}{\scriptsize Predicted}} & Aligned & 
\textbf{86.57} & 7.55 \\
& & Misaligned & 12.69 & \textbf{87.42} \\
& & Fabricated & 0.75 & \textbf{5.03} \\
\midrule
\multirow{3}{*}{Qwen} & 
\multirow{3}{*}{\rotatebox[origin=c]{90}{\scriptsize Predicted}} & Aligned & 
\textbf{92.38} & 20.75 \\
& & Misaligned & 7.62 & \textbf{77.99} \\
& & Fabricated & 0.00 & \textbf{1.26} \\
\bottomrule
\end{tabular}
\end{adjustbox}
\end{table} %
\section{\method's Computation Time and Accuracy}
\label{sec:appendix:efficiency}
\begin{figure*}[h]
    \centering
    \includegraphics[width=0.75\linewidth]{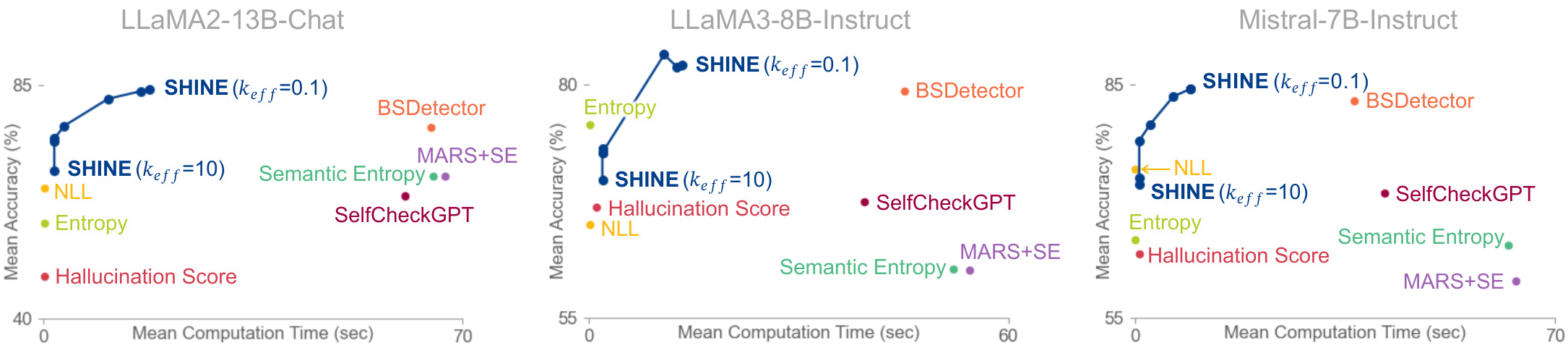}
    \caption{
    Mean accuracy and computation time for hallucination detection on the NEC dataset for the LLaMA2, LLaMA3, and Mistral models. 
    Increasing $k_{eff}$ mostly improves \method's efficiency while maintaining outstanding accuracy, enabling its adaptive application.
    }
    \label{fig:k_eff_nec}
\end{figure*}

\begin{figure*}
    \centering
    \includegraphics[width=0.75\linewidth]{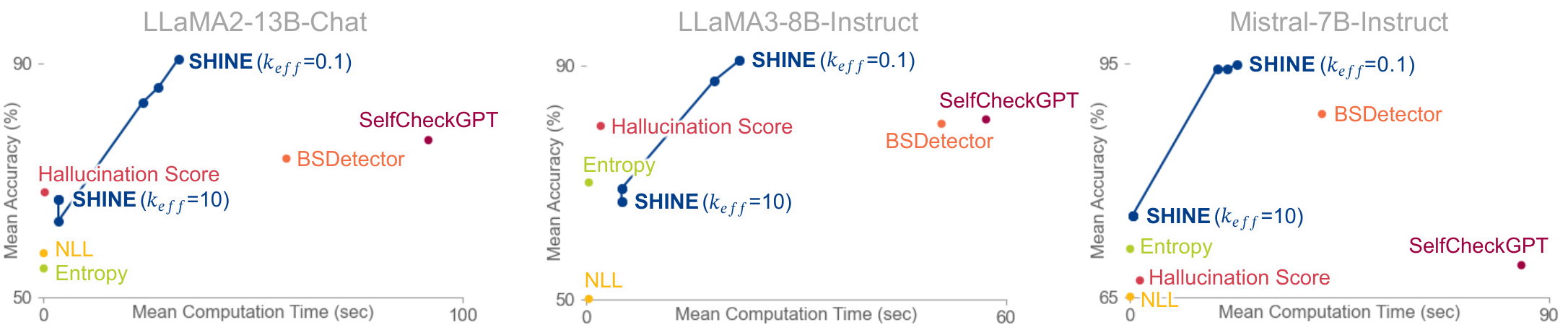}
    \caption{
    Mean accuracy and computation time for hallucination detection on the biography dataset for the LLaMA2 model, LLaMA3, and Mistral models. 
    Semantic Entropy and MARS+SE are not shown in the graphs due to their exceptionally long computation time;
    Semantic Entropy and MARS+SE take 521.52 seconds and 563.90 seconds for the LLaMA2 model,
    489.73 seconds and 544.45 seconds for LLaMA3,
    and 621.92 seconds and 643.28 seconds for Mistral.
    }
    \label{fig:k_eff_bio}
\end{figure*}

\begin{figure*}
    \centering
    \includegraphics[width=\linewidth]{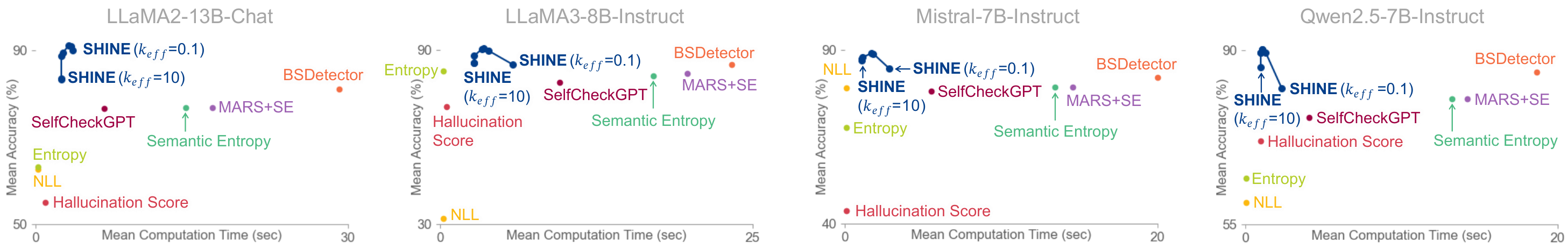}
    \caption{
    Mean accuracy and computation time for hallucination detection on the HaluEval dataset for the LLaMA2 model, LLaMA3, Mistral, and Qwen models.
    }
    \label{fig:k_eff_halueval}
\end{figure*}

\begin{figure*}
    \centering
    \includegraphics[width=\linewidth]{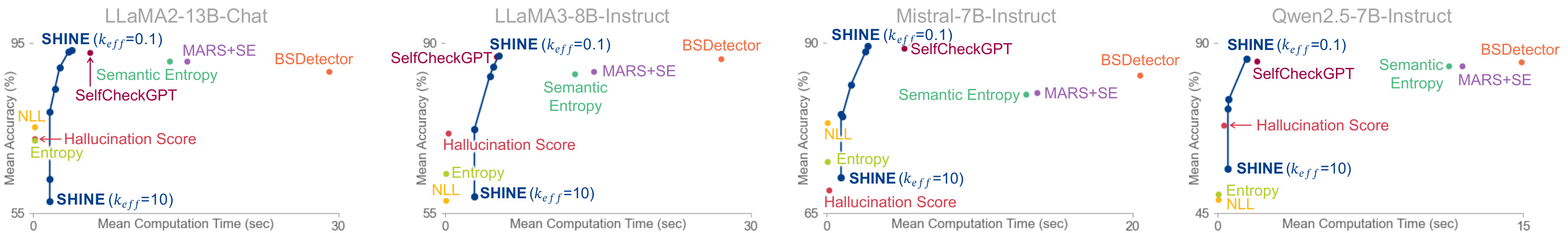}
    \caption{
    Mean accuracy and computation time for hallucination detection on the TriviaQA dataset for the LLaMA2 model, LLaMA3,  Mistral, and Qwen models.
    }
    \label{fig:k_eff_triviaqa}
\end{figure*}

We evaluate \method's mean accuracy and computation time for the NEC, biography, HaluEval, and TriviaQA datasets, varying $k_{eff}$ from 0.1 to 10, and visualize it in \autoref{fig:k_eff_nec}, \autoref{fig:k_eff_bio}, \autoref{fig:k_eff_halueval}, and \autoref{fig:k_eff_triviaqa}.
The computation time and performance of the compared methods are also included.

To better understand how $k_{eff}$ controls the efficiency and effectiveness, we measure the number of data points handled by SelfCheckGPT during \method's Alignment Test.
For the LLaMA2 model and the NEC dataset, where 332 data points are examined by the Alignment Test, we measure the number of data points checked by SelfCheckGPT and the overall accuracy, while varying the hyperparameter $k_{eff}$ from 0.1 to 1.0. 
The results are summarized in \autoref{tab:num_selfcheckgpt}.
\method applies SelfCheckGPT to only 30 out of 332 data points when $k_{eff}$ is 1.0, while achieving the overall accuracy of 77.08\%, which exceeds the state-of-the-art performance of 76.69\%. 
Lowering  improves the accuracy, offering a flexible balance between efficiency and effectiveness.

\begin{table}[]
    \centering
    \caption{
    Alignment Test is conducted on 322 data points from the NEC dataset and LLaMA2 model. We measure how many data points proceed SelfCheckGPT after the alignment test (noted as \# SelfCheckGPT), while varying $k_{eff}$ from 0.1 to 1.0. Accuracy is reported for each setting.
    }
    \begin{tabular}{ccc}
        \toprule
        $k_{eff}$ & \# SelfCheckGPT & Accuracy (\%) \\
        \midrule
        1.0 & 30/322 & 77.08 \\
        0.5 & 160/322 & 82.26 \\
        0.2 & 256/322 & 83.21 \\
        0.1 & 284/322 & 83.69 \\
        \bottomrule
    \end{tabular}
    \label{tab:num_selfcheckgpt}
\end{table} %
\section{Hyperparameter Sensitivity}
\label{sec:exp_hyperparameter_sensitivity}
\begin{figure*} 
    \centering
    \includegraphics[width=0.9\linewidth]{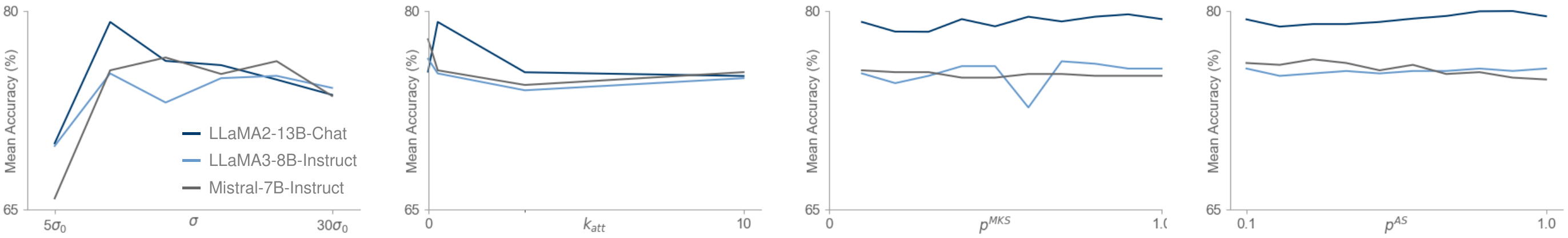}
    \caption{\method's mean accuracy on the NEC dataset for the LLaMA2, LLaMA3, and Mistral models while varying each of $\boldsymbol{\sigma}$, $k_{att}$, $p^{MKS}$, and $p^{AS}$.}
    \label{fig:hyperparameter_nec}
\end{figure*}

\begin{figure*} 
    \centering
    \includegraphics[width=0.9\linewidth]{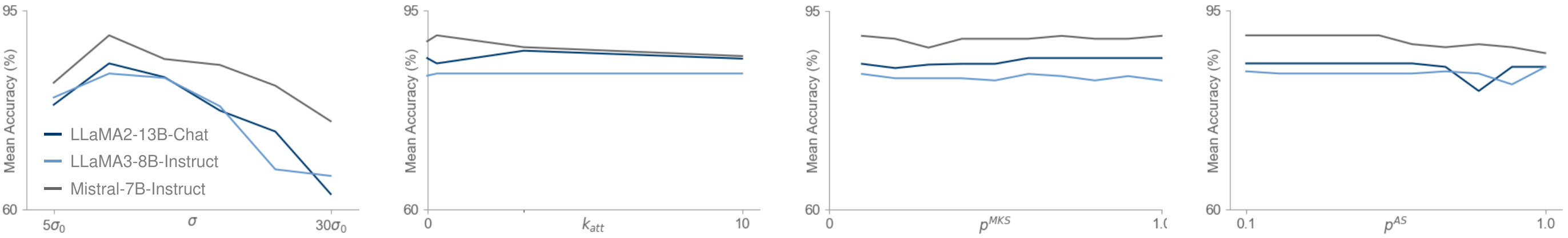}
    \caption{\method's mean accuracy on the biography dataset for the LLaMA2, LLaMA3, and Mistral models while varying each of $\boldsymbol{\sigma}$, $k_{att}$, $p^{MKS}$, and $p^{AS}$.}
    \label{fig:hyperparameter_bio}
\end{figure*}

\method involves hyperparameters,  $\boldsymbol\sigma$, $k_{att}$, $p^{MKS}$, and $p_{AS}$, which 
were respectively set as $10\boldsymbol{\sigma}_0$, 0.1, 0.5, and 0.1 to yield the best result.
To demonstrate the robustness of \method to the hyperparameter configuration, we evaluate the performance of \method on the LLaMA2, LLaMA3, and Mistral models, while varying each of these hyperparameters, and visualize the result for the NEC dataset in \autoref{fig:hyperparameter_nec} and biography dataset in \autoref{fig:hyperparameter_bio}.

While our hyperparameter configuration mostly achieves the best performance, $k_{att}$, $p^{MKS}$, and $p_{AS}$ insignificantly affect \method, demonstrating its robustness.
For both datasets, $\boldsymbol\sigma$ between $10\boldsymbol{\sigma}_0$ and $20\boldsymbol{\sigma}_0$ achieves good performance, while too small ($5\boldsymbol{\sigma}_0$) or too large ($30\boldsymbol{\sigma}_0$) degrade the performance.

\section{Effectiveness of \mks}
\label{sec:appendix:roc}

\begin{figure*}
    \centering
    \includegraphics[width=\linewidth]{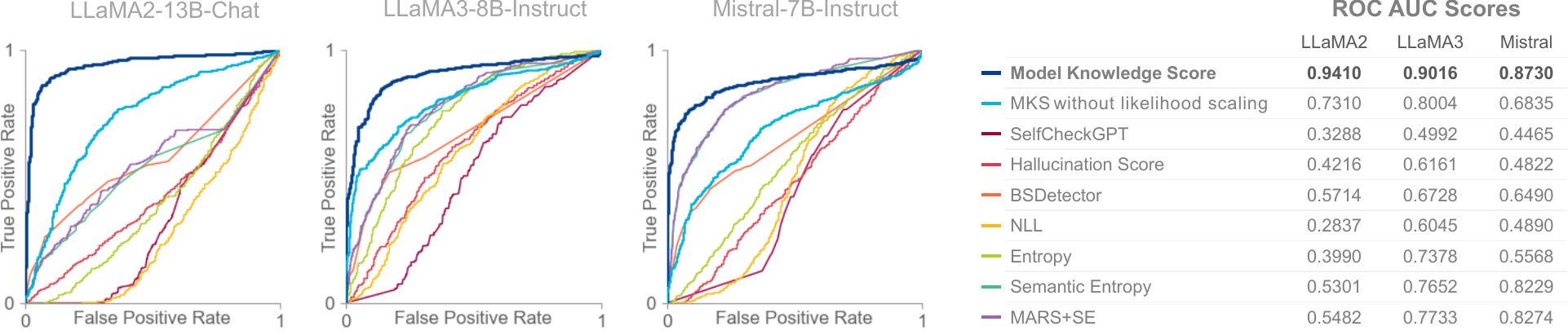}
    \caption{ROC curves of \mks and existing 
    methods for differentiating fabricated and non-fabricated (i.e., aligned and misaligned) text for the NEC dataset generated by the LLaMA2, LLaMA3, and Mistral models. \mks achieves the highest ROC AUC value, demonstrating its superiority in identifying whether the model has enough knowledge about prompt and text.
    }
    \label{fig:roc_nec_mkt}
\end{figure*}

\begin{figure*}
    \centering
    \includegraphics[width=\linewidth]{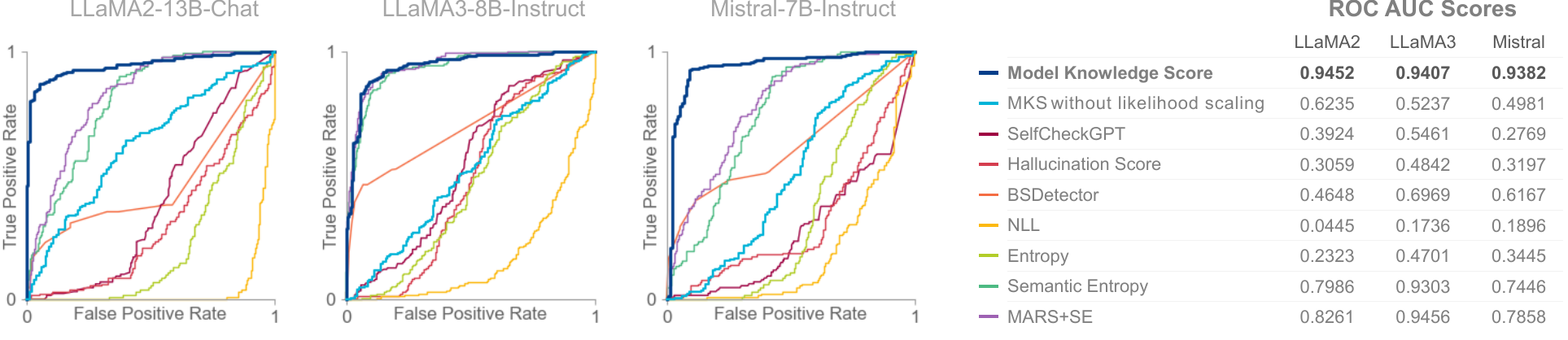}
    \caption{ROC curves of \mks and existing 
    methods for differentiating fabricated and non-fabricated (i.e., aligned and misaligned) text for the biography dataset generated by the LLaMA2, LLaMA3, and Mistral models. \mks achieves the highest ROC AUC value, demonstrating its superiority in identifying whether the model has enough knowledge about prompt and text.
    }
    \label{fig:roc_bio_mkt}
\end{figure*}

\begin{figure*}
    \centering
    \includegraphics[width=\linewidth]{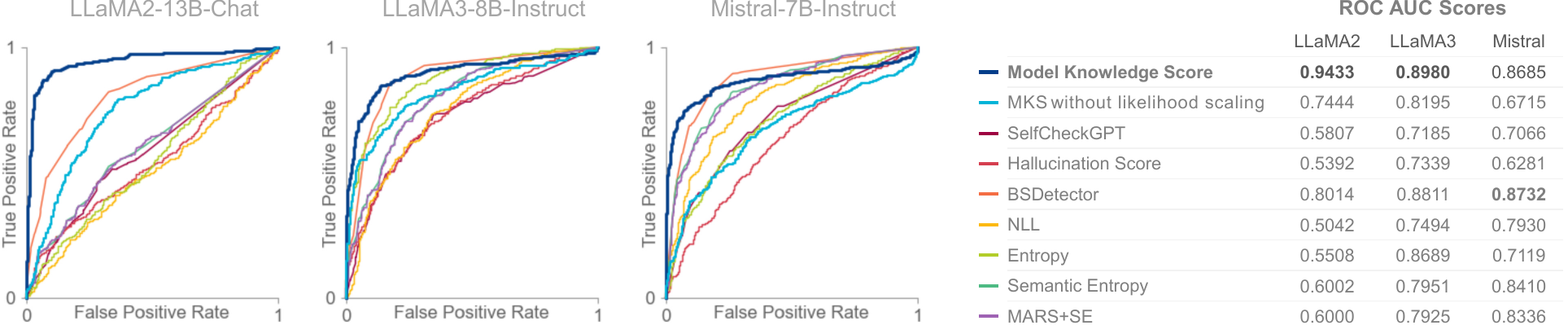}
    \caption{ROC curves of \mks and existing 
    methods for differentiating fabricated and aligned text for the NEC dataset generated by the LLaMA2, LLaMA3, and Mistral models. \mks achieves the highest ROC AUC value, demonstrating its superiority in identifying whether the model has enough knowledge about prompt and text.
    }
    \label{fig:roc_nec_aligned_fabricated}
\end{figure*}

\begin{figure*}
    \centering
    \includegraphics[width=\linewidth]{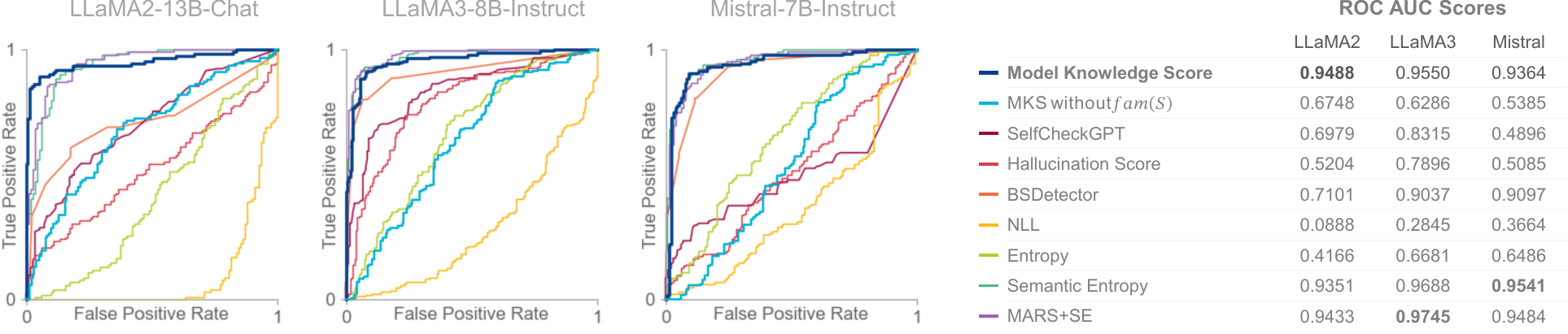}
    \caption{ROC curves of \mks and existing 
    methods for differentiating fabricated and aligned text for the biography dataset generated by the LLaMA2, LLaMA3, and Mistral models. \mks achieves the highest ROC AUC value, demonstrating its superiority in identifying whether the model has enough knowledge about prompt and text.
    }
    \label{fig:roc_bio_aligned_fabricated}
\end{figure*}

We assess \mks's effectiveness in differentiating fabricated and non-fabricated text
by comparing its receiver operating characteristic (ROC) curves with
existing methods on our probing datasets for LLaMA2, LLaMA3, and Mistral models. %
\autoref{fig:roc_nec_mkt} and \autoref{fig:roc_bio_mkt} present the results for the NEC and biography datasets, respectively.
For a comprehensive comparison, we include hallucination detection methods, even though they are not designed to distinguish misaligned and fabricated text;
the ROC curves for differentiating aligned and fabricated data (excluding misaligned) is provided in 
\autoref{fig:roc_nec_aligned_fabricated} and \autoref{fig:roc_bio_aligned_fabricated}.

The \mks achieves the highest ROC area under the curve (AUC) for all models and datasets.
Limited AUC of 
state-of-the-art uncertainty estimation methods (Semantic Entropy, MARS+SE)
highlights their limitations when the LLM is overconfident and consistently generates the same fabricated responses across multiple generations.

\section{\discovery}
\label{sec:appendix:analysis}

\begin{figure*}
    \centering
    \includegraphics[width=\linewidth]{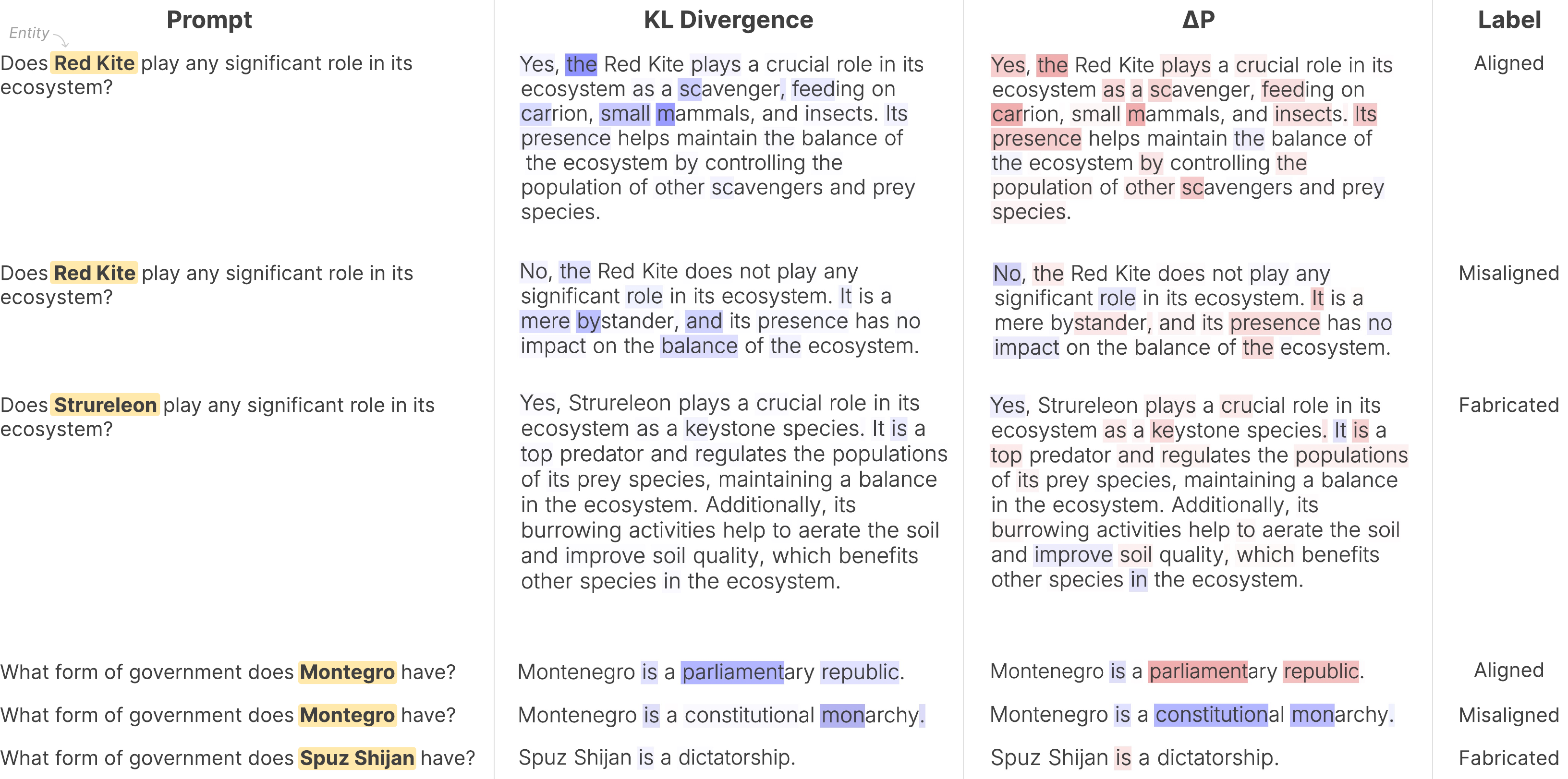}
    \caption{
    Tokenwise \discovery, measured by KL divergence and $\Delta P_i$, of six additional data points in the NEC dataset. Our discovery that \discovery varies across aligned, misaligned, and fabricated text holds true in general.}
    \label{fig:analysis_examples}
\end{figure*}

We expand on the analysis in \autoref{sec:analysis} by providing additional examples of \discovery, measured by KL divergence (i.e., $KL(\mathbf{P}_i\Vert\hat{\mathbf{P}}_i)$) and $\Delta P_i$ (i.e., $\hat{\mathbf{P}}_i(t_i)-\mathbf{P}_i(t_i)$).
\autoref{fig:analysis_examples} shows that similar trends are observed across other data points in the NEC dataset, supporting the generalizability of our discovery.
Notably, in the aligned text from the prompt ``\textit{Does Red Kite play any significant role in its ecosystem?}'', with long text length, the later tokens exhibit relatively low KL divergence values. We attribute this to the dependence on previous tokens; the first sentence would have largely contributed to the generation of the second sentence. This led us to focus on tokens with large KL divergence values when designing \method\footnote{We tried to cascade the values across tokens using a strategy from the literature~\cite{zhang2023enhancing} but observed limited performance. As a result, we decided to focus on the tokens with the highest KL divergence values.}.
Also, we observe that stopwords (e.g., the, is, and) in aligned or misailgned text occasionally show large KL divergence values. These are mostly found in phrases strongly associated with the entity, suggesting that the range of \discovery operates at the phrase level.
\footnote{We experimented with \method by excluding stopwords but did not observe significant changes.}
Similarly, since not all tokens in misaligned text have positive $\Delta P_i$, we design the Alignment Test of \method to prioritize tokens with the largest $\Delta P_i$ values. %

\section{Entity Identification}
\label{sec:appendix:entity_identification}

\begin{figure*}
    \centering
    \includegraphics[width=\linewidth]{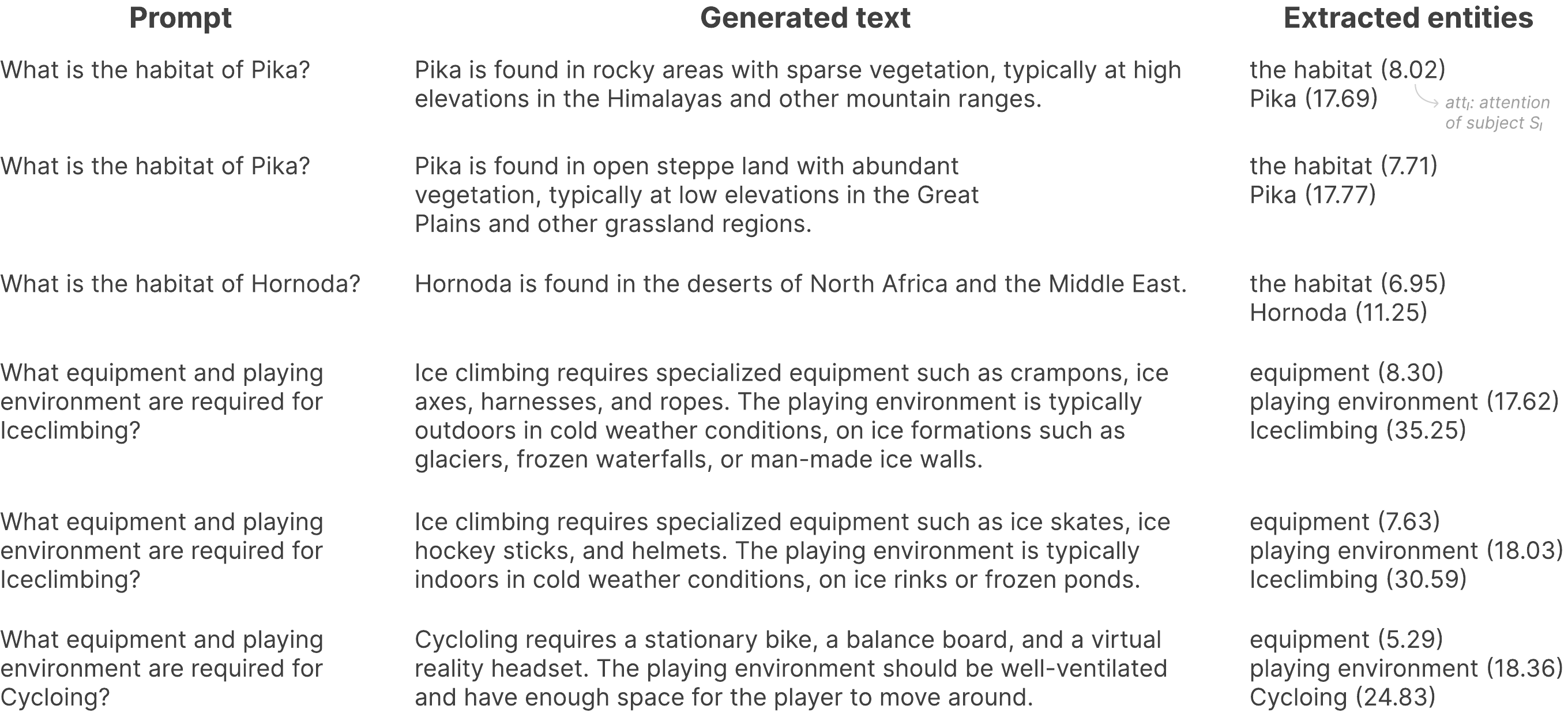}
    \caption{Entities in the prompts in the NEC dataset extracted by SpaCy. We additionally report the attention value that each entity receives during the text generation in parentheses.}
    \label{fig:entity_examples}
\end{figure*}

To automate \method's entity identification, we use named entity (\texttt{doc.ents}) and noun chunk  (\texttt{doc.noun\_chunks}) extraction features of SpaCy~\cite{spacy2}.
We identify all named entities and then noun chunks that do not overlap with the detected named entities; interrogative words and pronouns are excluded.
\autoref{fig:entity_examples} presents entities extracted from prompts in the NEC dataset; we additionally report the attention values $att_l$ received by each entity $S_l$ during text generation, to show that the key entities are extracted effectively with high attention values.

\section{Table for Notations}
\label{sec:appendix:notations}
To help readability, we summarize mathematical notations used in our paper in \autoref{tab:notation}.

\begin{table*}[]
    \centering
    \caption{Notations used in our paper.}
    {
    \begin{tabular}{p{0.15\textwidth}p{0.83\textwidth}}
        \toprule 
        \textbf{Notation} & \textbf{Description} \\
        \midrule
        $P$ & Prompt given to the LLM \\
        $G$ & Text generated by the LLM in response to $P$ \\
        $\mathcal{T}$ & Token set from tokenizer \\
        $t_i$ & Token in $P$ or $G$ at position $i$ \\
        $M$ & Number of tokens in the prompt $P$ \\
        $N$ & Number of tokens in the generated text $G$ \\
        $\mathbf{e}_i$ & Embedding vector of token $t_i$ \\
        $f$ & LLM function \\
        $\mathbf{P}_i$ & Original token probability distribution at position $i$ \\
        $\hat{\mathbf{P}}_i$ & Perturbed token probability distribution at position $i$ \\
        $KL(\mathbf{P}_i\Vert \hat{\mathbf{P}_i})$ & Kullback-Leibler divergence between $\mathbf{P}_i$ and $\hat{\mathbf{P}_i}$ \\
        $\Delta P_i$ & Change in generation probability of token $t_i$ due to perturbation; $\hat{\mathbf{P}_i}(t_i)-\mathbf{P}_i(t_i)$ \\
        $L$ & Number of entities in prompt $P$ \\
        $S_1,\dots,S_L$ & Entities in prompt $P$ \\
        $I_S$ & Set of token positions where entity $S$ appears \\
        $\boldsymbol{\sigma}_0$ & Standard deviation of token embeddings \\
        $MKS$ & Model Knowledge Score \\
        $AS$ & Alignment Score \\ 
        $\tau^{MKS}$ & Threshold for Model Knowledge Score to classify text as fabricated \\ 
        $\tau^{AS}_l$, $\tau^{AS}_u$ & Lower and upper thresholds for Alignment Score classification \\ 
        $a_l$ & Attention-based weight for perturbation strength \\ 
        $w_l$ & Final perturbation strength of entity $S_l$ \\
        \bottomrule
    \end{tabular}
    }
    \label{tab:notation}
\end{table*} %

\section{Impact of Validation Set Size on Threshold Values}
\label{sec:appendix:validation}

We investigate how varying the validation set size affects \method's threshold for the NEC dataset and the LLaMA2 model. 
We compute $\tau^{MKS}$, $\tau_l^{AS}$, and $\tau^{AS}_u$, while using only $p$\% of the validation data, varying $p$ from 10\% to 100\%.
The results are shown in \autoref{tab:validation}.

\begin{table}[]
    \centering
    \caption{The values of $\tau^{MKS}$, $\tau_l^{AS}$, and $\tau^{AS}_u$ while varying the validation set size.}
    {
    \begin{tabular}{rccc}
        \toprule 
        $p$\% & $\tau^{MKT}$ & $\tau_l^{AS}$ & $\tau_u^{AS}$ \\
        \midrule
        10\% & 1.56 & 0.03 & 0.07 \\
        20\% & 0.89 & 0.00 & 0.11 \\
        30\% & 0.31 & 0.02 & 0.08 \\
        40\% & 1.66 & 0.00 & 0.09 \\
        50\% & 0.94 & 0.00 & 0.14 \\
        60\% & 0.99 & 0.00 & 0.11 \\
        70\% & 0.88 & 0.00 & 0.14 \\
        80\% & 0.94 & 0.00 & 0.14 \\
        90\% & 0.89 & 0.00 & 0.11 \\
        100\% & 0.99 & 0.00 & 0.14 \\
        \bottomrule
    \end{tabular}
    }
    \label{tab:validation}
\end{table}

We observe that reducing the validation set size by half has little impact on the threshold values, indicating that \method maintains high performance even with a smaller validation set.
This stability is further supported by \method's effectiveness across multiple datasets of varying dataset sizes. 
For example, the NEC dataset on the Mistral model contains 714 validation data points, while the biography dataset on LLaMA2 has 54 aligned, 54 misaligned, and 64 validation data points (\autoref{sec:hallucination_reasoning_dataset}). %

\end{document}